\documentclass[pdflatex,sn-apa,oneside]{sn-jnl}

\usepackage{graphicx}
\usepackage{multirow}
\usepackage{amsmath,amssymb,amsfonts}
\usepackage{amsthm}
\usepackage{mathrsfs}
\usepackage[title]{appendix}
\usepackage{xcolor}
\usepackage{textcomp}
\usepackage{manyfoot}
\usepackage{booktabs}
\usepackage{algorithm}
\usepackage{algorithmicx}
\usepackage{algpseudocode}
\usepackage{listings}
\usepackage{natbib}
\usepackage{tikz}
\usepackage[T1]{fontenc}
\usepackage{newtxtext,newtxmath}

\theoremstyle{thmstyleone}

\usetikzlibrary{matrix, positioning, calc}

\theoremstyle{thmstyletwo}

\theoremstyle{thmstylethree}

\begin{document}

\title[Nonlinear Methods for Analyzing Pose in Behavioral Research]{Nonlinear Methods for Analyzing Pose in Behavioral Research}

%%=============================================================%%
%% GivenName	-> \fnm{Joergen W.}
%% Particle	-> \spfx{van der} -> surname prefix
%% FamilyName	-> \sur{Ploeg}
%% Suffix	-> \sfx{IV}
%% \author*[1,2]{\fnm{Joergen W.} \spfx{van der} \sur{Ploeg} 
%%  \sfx{IV}}\email{iauthor@gmail.com}
%%=============================================================%%

\author*[1,2,3]{\fnm{Carter} \sur{Sale}}\email{carter.sale@mq.edu.au}
\author[2,3,4]{\fnm{Margaret C.} \sur{Macpherson}}\email{cathy.macpherson@mq.edu.au}
\author[2,3]{\fnm{Gaurav} \sur{Patil}}\email{gaurav.patil@mq.edu.au}
\author[3,5,6]{\fnm{Kelly} \sur{Miles}}\email{kelly.miles@mq.edu.au}
\author[2,3]{\fnm{Rachel W.} \sur{Kallen}}\email{rachel.w.kallen@mq.edu.au}
\author[7]{\fnm{Sebastian} \sur{Wallot}}\email{sebastian.wallot@leuphana.de}
\author[2,3]{\fnm{Michael J.} \sur{Richardson}}\email{michael.j.richardson@mq.edu.au}

\affil[1]{\orgdiv{Modeling and Engineering Risk and Complexity}, \orgname{Scuola Superiore Meridionale}, \orgaddress{\city{Naples}, \country{Italy}}}

\affil*[2]{\orgdiv{School of Psychological Sciences}, \orgname{Macquarie University}, \orgaddress{\city{Sydney}, \country{Australia}}}

\affil[3]{\orgdiv{Performance and Expertise Research Centre}, \orgname{Macquarie University}, \orgaddress{\city{Sydney}, \country{Australia}}}

\affil[4]{\orgdiv{Lifespan Health and Wellbeing Research Centre}, \orgname{Macquarie University}, \orgaddress{\city{Sydney}, \country{Australia}}}

\affil[5]{\orgdiv{Macquarie University Hearing Research Centre}, \orgname{Macquarie University}, \orgaddress{\city{Sydney}, \country{Australia}}}

\affil[6]{\orgdiv{Department of Linguistics}, \orgname{Macquarie University}, \orgaddress{\city{Sydney}, \country{Australia}}}

\affil[7]{\orgdiv{Institute for Sustainability Psychology}, \orgname{Leuphana University of Lüneburg}, \orgaddress{\city{Lüneburg}, \country{Germany}}}

\abstract{Advances in markerless pose estimation have made it possible to capture detailed human movement in naturalistic settings using standard video, enabling new forms of behavioral analysis at scale. However, the high dimensionality, noise, and temporal complexity of pose data raise significant challenges for extracting meaningful patterns of coordination and behavioral change. This paper presents a general-purpose analysis pipeline for human pose data, designed to support both linear and nonlinear characterizations of movement across diverse experimental contexts. The pipeline combines principled preprocessing, dimensionality reduction, and recurrence-based time series analysis to quantify the temporal structure of movement dynamics. To illustrate the pipeline’s flexibility, we present three case studies spanning facial and full-body movement, 2D and 3D data, and individual versus multi-agent behavior. Together, these examples demonstrate how the same analytic workflow can be adapted to extract theoretically meaningful insights from complex pose time series.}


\maketitle

\section{Introduction}\label{sec1}

For decades, researchers have relied on laboratory-based motion capture systems to quantify human movement with high spatial precision \citep{winter2009biomechanics}. The sub-millimeter accuracy of marker-based systems, however, comes at the cost of high expense, restricted laboratory environments, and a marker application process that can constrain natural movement \citep{colyer2018review,seethapathi2019movement}. These constraints do more than limit where movement can be recorded: they shape what behaviors can be studied, often favoring simplified, highly controlled tasks at the expense of ecological validity. This trade-off has driven the search for more flexible and accessible approaches to movement measurement.

Recent advances in computer vision now enable detailed kinematic tracking from ordinary video via markerless pose estimation \citep{mathis2018deeplabcut,cao2019openpose,fang2022alphapose,roggio2024comprehensive}. These deep learning-based methods infer 2D or 3D coordinates for anatomical keypoints, such as shoulders, elbows, knees, and facial landmarks, by learning visual patterns from annotated training data \citep{lecun2015deep,toshev2014deeppose,zheng2023deep}.  Applied frame-by-frame to video, they yield landmark trajectories that support movement analysis without markers, substantially reducing the logistical and technical barriers associated with traditional motion capture.

As pose estimation methods have advanced, attention turned not just to the feasibility of their application but to how different algorithmic strategies handle complex, real-world scenarios. This distinction is relevant here because the type of pose-estimation model used upstream influences the kinds of tracking errors, occlusions, and identity-assignment problems that must be addressed downstream in preprocessing and analysis. Contemporary algorithms such as OpenPose, MediaPipe, RTM Pose, and DeepLabCut differ primarily in how they handle multi-person scenarios \citep{cao2019openpose,jiang2023rtmpose, lugaresi_mediapipe_2019,mathis2018deeplabcut}. Top-down methods first localize individuals via bounding boxes, then estimate keypoints within each box—an approach that excels for single subjects but struggles when bodies overlap or occlude one another \citep{khirodkar_multi-instance_2021}. Bottom-up methods reverse this logic, detecting all visible keypoints first and then clustering them into individuals, which scales better to crowded scenes at some cost to per-joint precision \citep{kresovic_bottom-up_2021,geng_bottom-up_2021}. Regardless of architecture, accuracy continues to improve through better training data, multi-view integration, and refined network designs.

Yet even with improving algorithms, what researchers can extract still depends heavily on the recording setup itself. This matters for the present work because acquisition geometry directly shapes the dimensionality of the data, the types of artifacts introduced, and the preprocessing required before meaningful dynamical analysis is possible. Single-camera 2D tracking is straightforward to deploy but sensitive to viewing angle: camera placement and participant orientation introduce variance that can dominate genuine movement signals \citep{seethapathi2019movement}. Three-dimensional reconstruction—achieved through multi-camera rigs, depth sensors, or stereo vision—mitigates these perspective effects and better captures the full geometry of movement, though at greater hardware cost and computational expense \citep{nogueira2025markerless}.

These differences lead to a central question: how accurate are markerless methods? A growing validation literature has compared them against gold-standard optical motion capture, revealing good agreement for global kinematic parameters despite reduced precision at individual joints, particularly under occlusion or rapid movement \citep{matsuda_validity_2024,evans_synchronised_2024,kosourikhina2022validation}. For behaviors requiring millisecond timing or sub-millimeter spatial resolution, specialized hardware remains necessary \citep{nystrom2025fundamentals,nakano2020evaluation}. In most behavioral science applications, however, markerless tracking provides sufficient fidelity, and these gains often justify the trade-offs \citep{pagnon2022pose2sim,lahkar2022accuracy}. Rather than replacing such systems, markerless approaches are best understood as a complementary methodology that trades absolute precision for gains in scalability, ecological validity, and accessibility.

A key consequence of these developments is the democratization of these tools, which has revolutionized data availability. This accessibility has opened the door to large-scale behavioral studies in naturalistic environments, where movement can be recorded across diverse contexts and analyzed at multiple spatial and temporal scales \citep{pagnon2022pose2sim,lahkar2022accuracy}. In doing so, it shifts the limiting factor in movement research from how data are collected to how they are analyzed. Recent work has also begun to formalize open-source pipelines that integrate pose estimation with dynamical analyses of behavioral interactions, highlighting the potential of computer-vision–based time-series approaches for scalable observational research \citep{bialek2025open}. Yet this accessibility introduces a distinctive set of analytical challenges. A single human pose typically comprises tens if not hundreds of tracked features, generating high-dimensional multivariate time series. Navigating this complexity is analytically non-trivial: without effective dimensionality reduction, the sheer volume of correlated features can dilute genuine effects and generate spurious associations, rendering conventional statistical methods unstable \citep{lone2024detecting,gloumakov2020dimensionality,halilaj2018machine}. These challenges are exacerbated by the fact that many standard analytical approaches were developed for low-dimensional, high-precision data and do not readily generalize to this setting.  Distinguishing signal from artifact is further complicated by the fact that natural movement variability is itself meaningful, reflecting functional fluctuations in timing and coordination \citep{bernstein1967coordination,latash2012bliss}. Pose estimation adds its own variability through tracking noise, missing keypoints, and frame-to-frame jitter \citep{simon20252d}. The challenge is to filter measurement artifacts while preserving the behavioral signal–a task requiring preprocessing pipelines that reduce noise and align data without erasing the structure of interest \citep{dixit2023contemporary,dindorf2024lab}. 

This justification is strengthened when we consider the kinds of questions typically asked in behavioral research. In many contexts, the scientific interest lies less in absolute spatial coordinates of movement and more in how movement evolves over time. Human behavior is often characterized not simply by where motion occurs, but by patterns of variability, regularity, and coupling within and between components of the system \citep{kelso_dynamic_1995,newell1986constraints}. For this reason, analyzing the temporal organization of movement becomes as critical as measuring its spatial precision  \citep{hausdorff1996fractal,delignieres2009fractal,diniz2011contemporary,marmelat2012strong}. However, extracting these dynamics requires more than standard linear approaches. While useful for quantifying magnitude, linear methods are blind to sequential organization and struggle with the non-stationary nature of pose data. 

To address these limitations, researchers have increasingly turned to methods from nonlinear dynamics \citep{riley1999recurrence,webber2005recurrence,shockley2005cross,nasim2019recurrence}. A powerful approach in this domain is Recurrence Quantification Analysis (RQA), which visualizes and quantifies the time-points when a system’s state returns to a previously observed state \citep{eckmann_recurrence_1987,zbilut_embeddings_1992}. Unlike linear measures, RQA makes no assumptions about linearity or stationary, making it well-suited for high-dimensional, noisy, and non-stationary pose data. 

Recurrence-based methods have already been widely applied to human motor coordination, postural control, gait dynamics, and interpersonal synchrony \citep{riley1999recurrence,shockley2005cross,ramdani_recurrence_2013}. As markerless pose estimation expands the accessibility and scale of movement measurement, RQA provides an established analytical framework for extracting temporal structure from these high-dimensional pose trajectories. Beyond movement, related nonlinear tools have also revealed cyclic and coordinated signatures in other complex signals, including brain activity \citep{lopes2021recurrence}, cardiovascular rhythms \citep{goldberger2002fractal}, and social coordination \citep{richardson2005looking}.

At its core, RQA is built on a simple intuition: systems often revisit similar states. In complex behavior, these returns to similar values can reflect underlying condition of the system in terms of organization, stability or strategies. To capture this, RQA transforms a time series into a higher-dimensional `state space' through a process called time delay embedding \citep{marwan_recurrence_2007}. This process creates a trajectory through the high-dimensional space, and its geometry is visualized in a recurrence plot—a matrix where a point is placed any time the system’s state at one moment is sufficiently close to its state at another \citep{eckmann_recurrence_1987}. The visual patterns that emerge are highly informative: diagonal lines indicate predictable, repeating sequences, while vertical or horizontal structures reflect periods where the system's dynamics persist or stabilize (Figure~\ref{fig:rqa_state_space}).

These metrics then allow structured patterns to be quantified \citep{webber_dynamical_1994,webber2005recurrence}. For instance, Determinism (DET) measures the proportion of recurrent points that form diagonal lines, acting as an index of predictability. Maximum line length ($L_{\text{max}}$) captures the longest sequence of sustained predictable behavior, reflecting stable coordination or persistent dynamical patterns. In contrast, Laminarity (LAM) captures the prevalence of vertical lines, reflecting the system's tendency to remain in a stable state. By quantifying these structural features, RQA offers a robust method for revealing temporal organization in behavioral data that often defies traditional analysis \citep{webber_dynamical_1994,marwan_recurrence-plot-based_2002,ramdani_recurrence_2013,crone2021synchronous}.

However, applying RQA to high-dimensional pose data is far from straightforward. Preprocessing, dimensionality reduction, parameter selection, and windowing choices can substantially reshape recurrence structure, potentially obscuring behavioral organization or amplifying measurement artifacts. Consequently, meaningful application of RQA requires a principled workflow that spans filtering, feature selection, and interpretation, while preserving the underlying movement dynamics \citep{marwan_recurrence_2007,wallot_multidimensional_2016}.

The diversity of pose-based research contexts further complicates matters. Movement data are collected across vastly different domains—from elite sports performance and rehabilitation \citep{fukushima2024potential}, to clinical gait assessment \citep{panconi2024deep}, to activities of daily living and workplace ergonomics \citep{walkling2025wearable}. This heterogeneity means that any analysis method must be both flexible and robust, capable of handling variations in sampling rates, tracking accuracy, movement types, and experimental constraints \citep{ji2023review}. Yet many existing pose-analysis approaches are narrowly tailored to specific modalities like sport \citep{noorbhai2025conceptual}. The absence of a broadly applicable solution highlights the need for a framework that can adapt to different datasets and research goals without requiring a complete redesign for each new application.

In this work, we present a general methodological framework for analyzing movement using pose data. This approach is implemented as a structured pipeline that integrates linear kinematic and recurrence analysis, bridging these gaps. We demonstrate its utility through three case studies illustrating robustness across contrasting experimental contexts, including 2D and 3D tracking, facial and full-body scales, and individual versus interpersonal dynamics.

\section{General Method}\label{sec:generalmethod}

The proposed pipeline treats movement not as a sequence of static postures, but as a continuous dynamical system \citep{kelso_dynamic_1995,Turvey1990coordination}. The workflow proceeds through four stages: (1) pose acquisition, (2) feature extraction, (3) preprocessing, and (4) analysis of the dynamics. This final stage operates at two complementary levels: linear kinematic metrics (e.g., velocity, displacement) are used to quantify movement magnitude, while Recurrence Quantification Analysis (RQA) is applied to characterize temporal organization–revealing coordination patterns, stability, and transitions invisible to amplitude-based measures.  

Rather than prescribing rigid protocols, the following sections provide decision frameworks and parameter guidelines designed to preserve this structure across diverse datasets. A step-by-step guide to the full workflow is provided in the Supplementary Materials, outlining the main decision points across acquisition, feature selection, preprocessing, and analysis.

\subsection{Pose Acquisition}

The pipeline starts with pose acquisition, where movement is recorded as coordinate trajectories of anatomical landmarks, independent of the specific capture modality. Whether derived from marker-based motion capture (e.g., Vicon, OptiTrack) or markerless estimation from video (e.g., OpenPose, MediaPipe, RTMPose, DeepLabCut), the required input is identical: frame-by-frame coordinate trajectories for anatomical landmarks. Readers seeking comprehensive reviews of markerless pose estimation technologies are referred to recent surveys \citep[e.g.,][]{roggio2024comprehensive,nogueira2025markerless,ji2023review,zheng2023deep}.

However, the validity of downstream analysis–particularly Recurrence Quantification Analysis (RQA)–imposes specific constraints on acquisition that differ from standard biomechanical approaches. While biomechanics prioritizes absolute spatial accuracy \citep{winter2009biomechanics}, dynamical analysis prioritizes temporal stability \citep{stergiou2011human}. Accordingly, the present framework is most appropriate in contexts where temporal consistency can be maintained, even if spatial precision is reduced. 

First, the sampling rate must be constant. Some consumer webcams and markerless algorithms produce variable frame rates, where the time ($\Delta t$) between frames fluctuates. Because time-delay embedding relies on a fixed temporal unit $\tau$, this jitter distorts the reconstruction of the phase space. If the acquisition hardware does not guarantee a fixed frame rate, timestamps must be recorded to allow for resampling during preprocessing.

Second, the recording configuration dictates both the dimensionality of the data and the nature of the artifacts. Single-camera 2D tracking is dependent on camera placement, where variance caused by viewing angle or participant orientation can be indistinguishable from genuine movement. Three-dimensional reconstruction mitigates these perspective effects, but at the cost of increased dimensionality and reconstruction noise (e.g., triangulation errors). The pipeline assumes that the input data contains both signal and these configuration-specific artifacts; consequently, the raw acquisition output is rarely suitable for direct analysis and must be conditioned through the specific preprocessing steps outlined below. Violations of these constraints do not simply degrade data quality, but introduce systematic distortions in the reconstructed dynamics that can lead to misleading or uninterpretable results.

\subsection{Feature Selection}

Pose estimation produces large numbers of highly correlated time series across landmarks. As such, the next step is to select the trajectories of interest. Directly analyzing high-dimensional trajectories risks redundancy and instability in downstream measures, yet reducing dimensionality too aggressively can obscure meaningful dynamics. The challenge is selecting a representation that preserves theoretically relevant aspects of movement while minimizing noise and redundancy.

The choice of features depends on the research question and the scale of movement under investigation. \textbf{Raw keypoint trajectories} $(x,y)$ or $(x,y,z)$ preserve anatomical specificity, enabling interpretation in terms of specific landmarks (e.g., pupil position, wrist displacement). This approach is valuable when localized features are of primary interest, though neighboring landmarks often move together, introducing redundancy. A useful variant collapses multi-dimensional coordinates into \textbf{magnitude vectors}–the Euclidean distance between successive frames.

This yields an orientation-agnostic measure of movement amplitude, reducing redundancy across axes while sacrificing directional information.

Alternatively, \textbf{derived kinematic metrics} compute bio-mechanically interpretable variables from landmark pairs or groups. Examples include aperture measures (vertical eyelid distance for blink amplitude, lip separation for mouth opening), rotational variables (head pitch, yaw, roll estimated from facial geometry), or relative displacements (distances between contralateral joints or from a landmark to a reference anchor). These signals can be differentiated to yield velocity (first derivative) and acceleration (second derivative) time series. Derived metrics offer strong interpretability and direct behavioral relevance–Case Study 1 uses blink amplitude as a proxy for attention and mouth aperture for speech activity–but are necessarily selective, capturing only a slice of available dynamics. They are most valuable when guided by clear theoretical rationale or domain knowledge.

In practice, no single representation is optimal for all pose datasets. A useful default is to begin with the lowest-dimensional representation that still preserves the behavior of theoretical interest. When a specific functional signal is the target, derived metrics are usually the best starting point because they provide direct interpretability and reduce redundancy. When overall movement intensity is of interest but directional information is not essential, magnitude vectors provide a robust and compact summary. Raw keypoint trajectories are most appropriate when anatomical specificity or fine-grained spatial structure must be preserved, but they should generally be reserved for cases where this additional detail is necessary, as they increase dimensionality and susceptibility to tracking noise. Thus, as a practical rule, we recommend starting with derived or magnitude-based features, and moving to full raw trajectories only when the scientific question requires landmark-level detail that simpler representations would obscure. The case studies below illustrate these choices in practice: Case 1 uses derived facial metrics, Case 2 analyzes regional keypoint clusters, and Case 3 applies magnitude vectors to end effectors of 3D data.

\subsection{Preprocessing}

Raw pose trajectories contain both genuine movement signals and systematic artifacts. Given this, before applying kinematic or recurrence analysis, data must be conditioned to isolate behavioral patterns from measurement noise. 

Preprocessing involves three main operations: handling missing data from tracking failures or occlusions, correcting for spurious variance introduced by camera geometry and participant positioning, and normalizing signals so that scale differences do not bias downstream measures. These steps are not a fixed recipe but a set of principles adapted to each dataset's characteristics. The guiding principle is to minimize nuisance variance while preserving the temporal structure that reveals coordination dynamics. 

\subsubsection{Handling Missing Data}

One of the first preprocessing challenges is handling missing data. Markerless pose estimation is inherently susceptible to missing data from occlusions, tracking failures, and low-confidence detections. Previous studies have typically applied linear or cubic interpolation across all gaps \citep{pagnon2022pose2sim}, but this approach can introduce substantial artifacts. Interpolation over long gaps alters the temporal structure of the signal–a particularly critical issue for recurrence analysis, which is sensitive to the sequential organization of states. Alternative approaches include probabilistic imputation \citep{kucherenko_neural_2018}, template-based completion \citep{liu_estimation_2006}, or simply excluding windows with excessive missingness. When RQA is not required, more flexible imputation methods may be appropriate. However, for recurrence-based analyses, interpolation length must be constrained to avoid fabricating dynamics that never occurred; in practice, extended gaps are better excluded than reconstructed. 

The constraint arises from how RQA reconstructs state space through time-delay embedding (detailed in Section~\ref{sec:rqa}). A one-dimensional signal $x(t)$ is transformed into vectors containing the current observation plus observations from $\tau$ time steps ago, $2\tau$ time steps ago, and so forth: $[x_t, x_{t-\tau}, x_{t-2\tau}, \ldots]$. The embedding delay $\tau$ captures meaningful temporal relationships, while the embedding dimension $m$ determines how many past observations are included.

When a gap of length $L$ is filled by interpolation, any reconstructed vector overlapping that gap will contain interpolated rather than observed values. Crucially, a gap affects not just the time points within it, but extends forward: vectors constructed $\tau$, $2\tau$, up to $(m-1)\tau$ steps after the gap ends still contain interpolated coordinates from within the gap. If the gap is long enough—specifically, longer than $(m-1)\tau$—some vectors will consist entirely of interpolated values–fully synthetic states that were not directly observed from the system. 

This reasoning suggests a conservative interpolation limit: $L_{\max} = (m-1)\tau$. Gaps shorter than this ensure every reconstructed vector contains at least one genuine observation. Beyond this threshold, the analysis begins operating on fabricated rather than measured dynamics.

For video-based pose estimation sample at 30-60 Hz, typical embedding parameters are $m=3-5$ and $\tau=10-25$ frames (estimation procedures described in Section~\ref{sec:rqa}). The conservative limit $L_{\text{max}}=(m-1)\tau$ therefore corresponds to gaps of approximately $0.5-2$ seconds–roughly the timescale of discrete behavioral events such as blinks, gestures, and postural shifts. Within this range, interpolation has minimal impact on recurrence structure; beyond it, phase-space distortions increase sharply. 

\begin{figure}
    \centering
    \includegraphics[width=\linewidth]{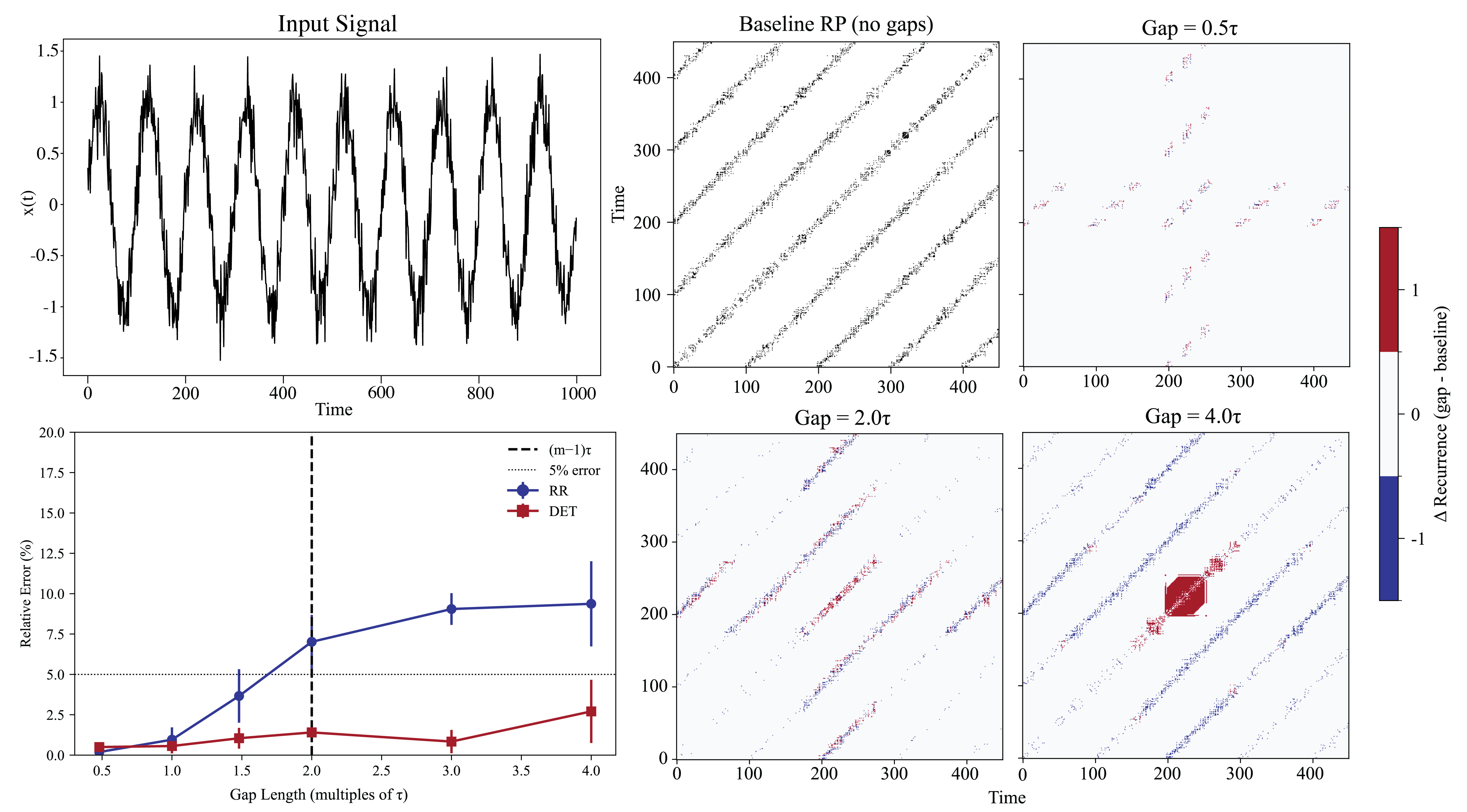}
    \caption{Effects of gap length on RQA stability and recurrence structure. Top Left: Raw time series showing original signal used. Bottom Left: Relative error in Recurrence Rate (RR) and Determinism (DET) as a function of the embedding delay $\tau$. Curves show the mean error across 30 trials, with error bars indicating standard deviation. The dashed line marks the theoretical threshold $(m-1)\tau$, beyond which embedding vectors begin to span the gap; the dotted line marks a 5\% error reference. RR error increases markedly once gaps exceed $(m-1)\tau$, while DET remains comparatively stable. Right: Recurrence plots illustrating how interpolation affects the underlying recurrence structure. The top-left panel shows the baseline RP (no gaps). Remaining panels show discrete difference maps (gaps-baseline) for centered gaps of 0.5$\tau$, 2$\tau$, and 4$\tau$. Colors denote recurrence changes: blue = lost recurrence, white = unchanged, red = gained recurrence. Small gaps introduce minimal distortions, whereas larger gaps generate structured artifacts, including diagonal streaks and localized regions of false recurrence.}
    \label{fig:interpolation}
\end{figure}

We validated this recommendation through controlled simulation using a noisy sine wave (signal-to-noise ratio 5:1), chosen as a canonical signal with known temporal structure that allows distortion to be directly quantified. This signal captures key characteristics of pose data: quasi-periodic structure from rhythmic movements combined with realistic measurement noise from tracking errors and natural movement variability (see Supplementary Materials). The signal was sampled at 100 Hz with period $T=1$ second. Embedding parameters were estimated using Average Mutual Information and False Nearest Neighbors analyses (further details provided in Section~\ref{sec:rqa}), yielding $m=3$ and $\tau=25$ samples–corresponding to one-quarter cycle of the dominant frequency. The theoretical threshold is thus $(m-1)\tau=50$ samples (0.5 seconds). 

We introduced gaps of varying length ($0.5\tau$ to $4\tau$), applied linear interpolation, and quantified deviations in recurrence metrics from ground truth values computed on gap-free data. Each gap length was tested across 30 trials with randomly positioned gaps (Figure~\ref{fig:interpolation}).

Results confirmed sensitivity to gaps exceeding the theoretical threshold. Recurrence rate—reflecting the overall density of state revisitations—increased from 0.2\% error at $0.5\tau$ to 7.0\% at the $(m-1)\tau$ threshold, reaching 9.4\% at $4\tau$ (Figure~\ref{fig:interpolation}A). Determinism—quantifying predictable sequential structure—remained more stable, showing only 1.4\% error at $2\tau$ and 2.7\% at $4\tau$. This differential sensitivity reflects that interpolation primarily inflates recurrence density through artificial smoothness rather than creating systematic sequential structure. The $(m-1)\tau$ threshold clearly demarcates a transition: recurrence rate remains below 4\% for gaps at or below this limit but approaches 10\% beyond it.

Visual inspection of recurrence plot difference maps confirmed these quantitative findings (Figure~\ref{fig:interpolation}B). Maps showing recurrences gained (red) or lost (blue) relative to ground truth revealed minimal artifacts at $0.5\tau$, subtle alterations at $2\tau$ coinciding with the theoretical threshold, and substantial structural changes at $4\tau$ indicating widespread phase-space distortion.

Based on these findings, we recommend $L_{\max} = (m-1)\tau$ as a conservative interpolation limit. This threshold maintains recurrence metric errors below 10\% for signals with realistic noise characteristics. The differential sensitivity of RQA metrics provides additional guidance when working with borderline cases: determinism proved more robust than recurrence rate to interpolation artifacts within the conservative range. This pattern suggests that if circumstances require interpolating gaps approaching or slightly exceeding $(m-1)\tau$, analyses should prioritize determinism, entropy, and line-length metrics over raw recurrence rate. However, we emphasize that staying within the $(m-1)\tau$ limit remains the safest approach, as all metrics show accelerating error beyond this threshold.

These limits represent conservative estimates validated under controlled conditions. The $(m-1)\tau$ threshold provides a robust starting point applicable across contexts without requiring signal-specific tuning, though researchers may verify robustness by comparing results with and without borderline windows for their specific datasets.

\subsubsection{Spatial Alignment}

Beyond missing data, preprocessing must also address spatial inconsistencies introduced by variations in recording set-up or participant positioning. These differences in absolute position, orientation, or scale across a dataset can introduce spurious variance that will be faithfully encoded in recurrence plots. Best practice is to minimize these discrepancies at the time of data collection (e.g., by fixing camera angles, distances, recording participant height). However, when this is not feasible, post-hoc spatial alignment becomes essential to ensure that the dynamics extracted from pose data reflect genuine movement, rather than differences in recording geometry \citep{vox2018preprocessing}.

Without alignment, differences in camera angle or participant orientation may cause the same action (e.g., raising a hand) to be processed as horizontal movement in one participant and vertical movement in another. Likewise, variation in scale (for instance, due to sitting closer or farther from the camera) can lead one participant’s movements to appear artificially larger in magnitude than another’s. For downstream analyses, these geometric differences distort recurrence structure by introducing variation unrelated to movement. 

One solution is to perform spatial alignment using a Procrustes method \citep{gower1975generalized,goodall1991procrustes}. This provides a principled way to remove differences in position, orientation, and scale by finding the optimal transformation that maps each participant’s pose configuration onto a shared template. In the classic formulation, this transformation is obtained via a singular value decomposition (SVD) of the cross-covariance matrix between the observed pose and the template. Formally, if \(X \in \mathbb{R}^{n \times d}\) denotes the observed pose configuration and \(T \in \mathbb{R}^{n \times d}\) the template, Procrustes alignment seeks the transformed configuration
\[
\hat{X} = sXR + \mathbf{1}t^\top
\]
that minimizes
\[
\min_{s,R,t} \; \|T - (sXR + \mathbf{1}t^\top)\|_F^2,
\]
where \(s\) is a scalar scale factor, \(R\) is an orthogonal rotation matrix, \(t\) is a translation vector, and \(\|\cdot\|_F\) denotes the Frobenius norm. This process comprises:
\begin{enumerate}
    \item \textbf{Translation:} subtracting the centroid of each frame so that all poses are compared relative to a common origin.
    \item \textbf{Scale:} estimating the uniform scalar that best matches the overall size of the participant’s keypoint configuration to the template.
    \item \textbf{Rotation:} finding the orthogonal rotation that minimizes the squared distance between the participant’s keypoint configuration and the template.
\end{enumerate}

\begin{figure}
    \centering
    \includegraphics[width=\linewidth]{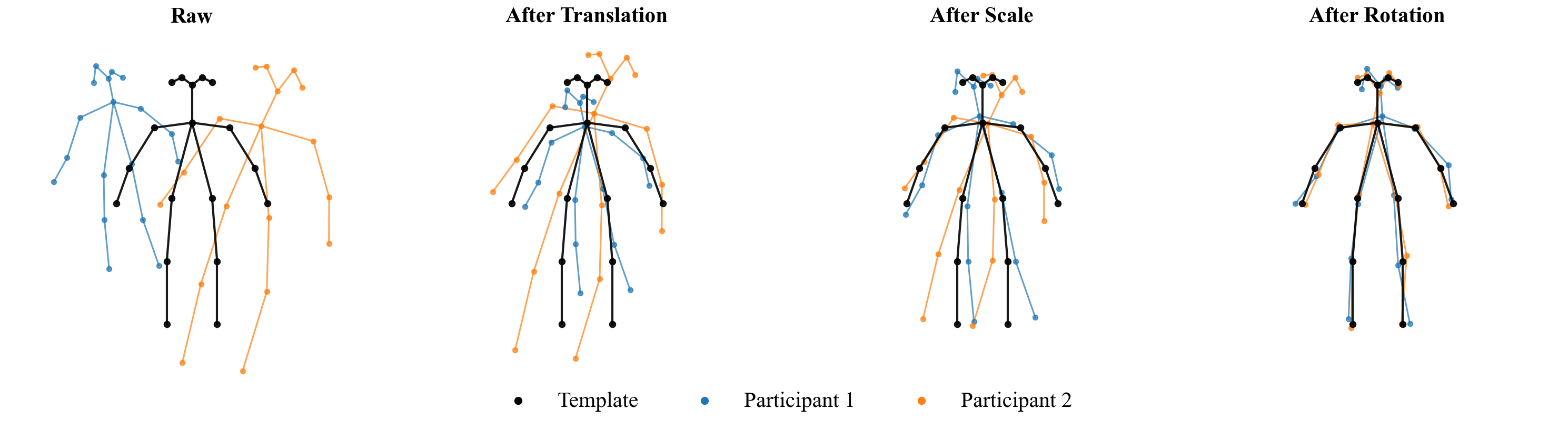}
    \caption{Procrustes alignment of 2D pose data. Synthetic pose skeletons illustrate the three stages of Procrustes alignment. (A) Raw: Two participants with different positions, scales, and orientations relative to the template (black). (B) After Translation: Poses are centered to the template centroid. (C) After Scale: Poses are normalized to match the template size. (D) After Rotation: Poses are rotated to align with the template orientation. This alignment procedure removes between-participant differences in position, body size, and camera angle, enabling direct comparison of pose dynamics.}
    \label{fig:procrustes}
\end{figure}

Once these parameters are estimated, each frame is transformed by applying the scale, rotation, and translation to all keypoint coordinates. In practical terms, the participant’s pose is: a) re-centered, b) rotated into the same global orientation as the template, and c) uniformly resized. The resulting pose is therefore expressed in a shared coordinate system that is comparable across participants and trials, while preserving the relative configuration of keypoints within each frame as well as moment-to-moment changes across frames. When performing Procrustes, it is useful to visualize the template and inspect a small number of raw versus aligned frames to verify that the transformation is working as expected (Figure~\ref{fig:procrustes}).

Notably, this approach requires researchers to either supply or construct a template to which all pose data are aligned. A template can be created by averaging keypoint positions across the dataset (yielding a global template; see each case study below), by selecting a single well-captured reference frame, or by generating a canonical pose using synthetic data. Optionally, researchers may choose to symmetrize the global template---enforcing a consistent midline and forward-facing orientation---to reduce dataset-level asymmetries. However, this may also suppress meaningful individual asymmetries (such as dominant hand usage) and is therefore best applied cautiously.

\subsubsection{Amplitude Normalization}
The last step in preprocessing is normalization to ensure comparability across participants, sessions, and features. Recurrence analysis relies on pairwise distances in reconstructed state space. As such, differences in scale directly bias both the recurrence threshold and the resulting metrics \citep{eckmann_recurrence_1987,marwan_recurrence_2007}. A variable with larger amplitude variation will dominate distance calculations unless scaled appropriately, even if its underlying dynamics are no more structured than those of other signals. The goal of normalization is therefore not to homogenize the data indiscriminately, but to choose a scaling convention that preserves the aspects of movement relevant to the analysis while minimizing nuisance variability.

Two common approaches are useful. Z-scoring (subtracting the mean and dividing by the standard deviation) removes differences in mean and variance, placing all signals on a common variance scale. This is particularly appropriate when amplitude is treated as nuisance variance and the focus is on recurrence structure itself, for example when comparing recurrence rate across participants or conditions. 

Unit-interval scaling (rescaling each signal to the [0,1] range) also removes absolute scale differences between signals, but preserves the proportional structure of fluctuations within each time series. In other words, it retains how movement varies relative to that signal’s own dynamic range, while eliminating between-series magnitude differences. This approach can be useful for derived metrics with natural bounds, where the shape of the trajectory within its envelope is theoretically meaningful, but absolute amplitude differences are not the primary target of analysis.

The scope and timing of normalization depend on the intended analysis. For short recordings, normalization parameters are typically computed from the entire trial. For longer recordings where movement patterns shift over time, normalization can be applied globally—preserving amplitude variation across time—or locally within analysis windows to focus on temporal structure independent of global trends. These decisions interact with how analysis parameters are set and should be made in light of specific analytic goals, as discussed in Sections \ref{sec:linear_analysis} and \ref{sec:rqa}.

%The timing of normalization also matters. A general rule is to normalize signals prior to segmentation into analysis windows. Doing so preserves variance across windows while ensuring that thresholds are not recalibrated at each step. Window-level normalization can be tempting, but it effectively erases meaningful differences in variance across windows and risks biasing recurrence metrics. The caveat is that normalization should only be applied to data that have already been adequately cleaned: if outliers or long missing segments remain, they will distort the scaling and propagate artifacts. As such, preprocessing and missing-data handling should be finalized before normalization is performed.

Detrending may be necessary when pose signals contain slow drift (e.g., gradual postural changes in webcam recordings). Removing low-frequency trends or high-pass filtering can prevent recurrence measures from being dominated by these shifts, but filtering must be applied with care, since it can itself reshape the temporal structure. Researchers should therefore select a normalization scheme that matches their analytic goals: use z-scoring when amplitude is not of theoretical interest, unit-interval scaling when relative magnitude matters, and apply detrending only when drift is clearly a nuisance.  

In practice, a useful default for recurrence-based analyses is to normalize signals after cleaning and interpolation but before recurrence estimation, using trial-level z-scoring when amplitude is not itself of theoretical interest. Unit-interval scaling is more appropriate when the relative shape of fluctuations within a bounded signal matters more than absolute variance. Global normalization should generally be preferred over window-wise normalization when comparisons across time are important, since local normalization can remove meaningful between-window amplitude differences. Detrending should be applied only when slow drift is clearly artifactual rather than behaviorally meaningful. Normalization should only be performed on adequately cleaned data, as outliers or long missing segments will distort scaling parameters and propagate artifacts.

\subsection{Linear Kinematic Analysis}\label{sec:linear_analysis}

Once pose trajectories have been cleaned, aligned, and represented in an appropriate feature space, the next step is to quantify how movement varies across tasks, conditions, or time. 

Linear kinematic analyses provide a first pass at this structure. These methods are appropriate when the primary research question concerns the magnitude, rate, or variability of movement, rather than its underlying temporal organization or coordination dynamics. They treat the signal as a sequence of positions evolving in time and summarize that evolution in terms of displacement, velocity, acceleration, and patterns of covariation across landmarks. These methods are ``linear'' in the sense that they operate on the raw or linearly transformed coordinates and do not explicitly model nonlinear state-space structure.

In the pipeline, linear analyses serve three roles. First, amplitude-based metrics quantify overall movement magnitude and variability, establishing whether experimental manipulations change how much participants move, how fast they move, or how abruptly they change direction. Second, multivariate methods such as principal components analysis (PCA) characterize dominant modes of coordinated motion across landmarks, revealing low-dimensional spatial structure in the movement patterns. Third, these linear summaries provide context for interpreting recurrence-based analyses: they help distinguish changes in temporal organization from differences that can be attributed simply to overall activity levels or shifts in dominant movement modes.

The interaction with preprocessing and normalization is direct. Amplitude-based metrics are usually computed on signals that retain physically meaningful units (e.g., pixels mapped to normalized screen coordinates or metric 3D coordinates), so that differences in mean displacement or velocity can be interpreted as genuine changes in movement magnitude. By contrast, PCA is typically applied to standardized trajectories (e.g., z-scored across time or across landmarks) to prevent high-variance coordinates from dominating the decomposition. Both types of linear analysis can be applied on entire trials or on windowed segments. Windowing allows linear summaries to track how movement magnitude or dominant postural modes evolve over time, rather than reducing each trial to a single global value.

The subsections below outline the amplitude-based metrics used in the pipeline and then describe PCA as a complementary tool for capturing coordinated multijoint structure.

\subsubsection{Amplitude-Based Metrics}

Amplitude-based metrics quantify the size and rate of movement directly from pose-derived time series. Starting from landmark trajectories or derived scalar features, velocity and acceleration are obtained through standard finite-difference operations on the position or velocity signals. The same procedures apply to any one-dimensional feature (e.g., apertures or inter-landmark distances), allowing kinematic derivatives to be computed uniformly across raw and derived measures.

These kinematic series are summarized using simple statistics that capture central tendency and variability. Common choices include the mean, standard deviation, maximum, and root-mean-square (RMS) of displacement, velocity, and acceleration over a given analysis interval. RMS measures are particularly useful, as they weight larger excursions more heavily and provide a compact index of the ``energetic'' level of movement \citep{moe1998new}. When multiple landmarks belong to a common anatomical or functional region (e.g., both wrists, or a set of facial points), per-landmark metrics can be averaged or otherwise pooled to produce region-level summaries. Global indices of whole-body movement can be obtained by aggregating across a defined subset of keypoints.

Windowing and normalization decisions follow directly from analytic goals. When absolute magnitude is meaningful, trajectories should retain a consistent spatial scale—either physical units or normalized coordinates that preserve amplitude differences. Detrending is preferable to per-window z-scoring in such cases, as the latter removes meaningful between-window variance \citep{pataky2024trial,wannop2012normalization}. When amplitude differences are considered nuisance variance, normalization may be applied, provided it is consistent across participants or conditions \citep{pinzone2016comprehensive, hof1996scaling}. 

Amplitude metrics can be computed within the same sliding windows used for recurrence analysis, enabling time-resolved comparisons between movement magnitude and temporal organization \citep{collins1993open}. Although these linear measures provide essential context, they do not capture the temporal structure of variability. As established in nonlinear movement science, two signals may exhibit identical amplitude statistics yet differ fundamentally in their dynamical organization \citep{stergiou2011human, kkedziorek2020nonlinear}. This distinction motivates the complementary use of recurrence-based methods, which explicitly quantify temporal structure when the goal is to characterize how movement unfolds over time.

%As a result, linear kinematic metrics are appropriate for questions about movement magnitude, variability, and spatial coordination, but are insufficient when the goal/research question is to characterise how movement unfolds over time. In such cases, methods that explicitely quantify temporal structure - such as recurrence-based analyses are required. 

\subsubsection{Principal Components Analysis (PCA)}

As mentioned previously, pose data involves many highly correlated trajectories. Neighboring keypoints tend to move together, global postural shifts can dominate variance, and redundant dimensions obscure the structure of interest. Principal component analysis (PCA) provides a principled means of identifying the dominant modes of coordinated movement by decomposing the covariance structure of the trajectories into orthogonal components. Each component captures a direction of maximal shared variance, and each principal movement reflects a coherent pattern of spatial co-variation across landmarks. These principal movements can be visualized directly as coordinated spatial patterns, an approach demonstrated in the Mirror Game case study (see Section~\ref{sec:mg_pca}). Formally, the scores $y_i = X v_i$ represent projections of the original data $X$ onto eigenvectors $v_i$ ordered by explained variance. Although this formulation is linear, it often reveals meaningful biomechanical or behavioral synergies. In hand posture control, for example, \citet{santello1998postural} demonstrated that just two principal components can account for over 80\% of variance across 15 joint angles during grasping, establishing PCA as a tool for identifying low-dimensional coordination patterns. \citet{daffertshofer2004pca} provide a comprehensive tutorial on using PCA to study coordination and variability in motor control, explaining both its role as a feature extractor and as a data-driven filter for separating invariant coordination structure from noise.

In practice, PCA is most useful when the goal is to identify dominant \textit{spatial} modes of coordinated movement or to reduce redundancy across highly correlated landmarks. It is less appropriate when the primary goal is to preserve the full temporal geometry of the signal for recurrence-based analysis, since the projection can alter the state-space structure on which recurrence measures depend. As a general rule, we recommend PCA as a tool for summarizing spatial organization and diagnosing preprocessing quality, rather than as a default preprocessing step prior to RQA.

The level at which PCA is performed has important implications for both interpretability and comparability. Performing PCA at the individual level captures the covariance structure specific to each participant and often yields components that align closely with their idiosyncratic movement patterns. This can improve anatomical interpretability but makes cross-participant comparison difficult, since the component axes differ from person to person. By contrast, group-level PCA—where data from all participants are pooled into a single decomposition—produces a common basis onto which every individual can be projected. This facilitates direct comparison across conditions or dyads but may obscure meaningful person-specific structure if movement styles differ substantially. The choice between individual- and group-level PCA therefore depends on the analytic objective: individual-level decompositions provide participant-specific insight, whereas group-level PCA supports comparisons within a shared coordinate system. Although several approaches have been proposed for reconciling group-level and individual-level structure in movement datasets (e.g., extensions of functional \citep{coffey2011common} or multilevel PCA \citep{di2009multilevel}), these methods address questions beyond the scope of the present pipeline. For the purposes of pose-based behavioral analysis, it is generally sufficient to treat individual- and group-level PCA as complementary perspectives on coordinated movement.

Normalization decisions interact closely with these considerations. Because PCA is sensitive to differences in variance across coordinates, the trajectories must be placed on a scale that matches the intended interpretation. When absolute displacement magnitude is meaningful, centering without further scaling preserves differences in movement amplitude. When the aim is to identify coordinated structure independent of scale, z-scoring each coordinate ensures that high-variance landmarks do not dominate the decomposition. Alternative normalization schemes have been proposed in the movement-science literature—including approaches that retain amplitude information in separate components or that normalize coordinates by body size—but these decisions depend on the specific research context rather than any intrinsic requirement of PCA \citep{deluzio2007biomechanical,federolf2016novel}. What is essential for pose-based analysis is that whatever normalization is chosen must be applied consistently across all data contributing to a given PCA space; inconsistent scaling alters the covariance structure and undermines the coherence of the resulting components. Consistent with this, recent sensitivity analyses show that different preprocessing choices do not change the total variance identified but can affect how variance is distributed across components, influencing their subsequent interpretation \citep{armstrong2024sensitivity}.

A further complexity arises when PCA is used in conjunction with recurrence analysis. Although PCA produces orthogonal components with respect to the covariance structure used to compute the decomposition (e.g., at the group level), this orthogonality does not guarantee that projected trajectories from an individual will remain uncorrelated. If $V$ denotes the group-level loadings and $\Sigma_i$ the covariance matrix of an individual’s data, then the covariance of that individual’s projected scores is $\mathrm{Cov}(S_i) = V^\top \Sigma_i V$, which is generally non-diagonal unless $\Sigma_i$ matches the group covariance exactly \citep{jolliffe2011principal}. In practical terms, this means that principal component scores may exhibit residual correlations at the individual level.

Because recurrence analysis relies on Euclidean distances in reconstructed state space, such induced correlations alter the geometry of that space. Correlated components effectively stretch or compress certain directions, introducing anisotropy into the distance metric and thereby inflating or suppressing recurrence density independently of the underlying temporal dynamics. For auto-recurrence, this may constrain trajectories to an apparently lower-dimensional manifold; for cross-recurrence, it may bias which aspects of coordination appear most salient. This issue has been noted in applied work using PCA prior to recurrence analysis. For example, \citet{straiotto_recurrence_2023} applied PCA to 45-dimensional center-of-mass trajectories in a martial arts task to obtain low-dimensional score time series suitable for RQA, but emphasized that PCA itself \textit{``does not access the temporal structure of the components''}, requiring subsequent nonlinear analysis to recover dynamical information.

These issues reflect a broader point established in movement neuroscience and physiological signal analysis: linear projections such as PCA often fail to preserve the nonlinear manifold structure that underlies many biological time series. For example, neural population activity has been shown to evolve on curved, low-dimensional manifolds that PCA only partially captures \citep{fortunato2024nonlinear}, and similar observations have been made in cardiac dynamics, where recurrence-relevant features reside in nonlinear subspaces that linear PCA distorts \citep{erem2012manifold}. Work on dynamical systems likewise argues that standard PCA may fail to capture critical nonlinear characteristics, especially when data lengths are short, proposing symplectic PCA as a more appropriate variant for such settings \citep{lei2011symplectic}. Collectively, these results do not imply that PCA is unsuitable for movement data, but they clarify why PCA alters the geometry in ways that are not neutral for recurrence analyses.

For these reasons, PCA is more naturally treated as a tool for examining spatial structure rather than as a preprocessing step for RQA. Its most appropriate role in a recurrence-based workflow is downstream: once RQA summary measures (e.g., recurrence rate, determinism, entropy) have been computed, PCA can be applied to these correlated features to reveal higher-order organization without altering the underlying temporal dynamics \citep{zbilut1998recurrence,orlando2021recurrence}.

Beyond its analytic utility, PCA also functions as a diagnostic tool for evaluating preprocessing quality. When spatial alignment or centering is inadequate, the leading components often capture global translation, slow drift, or other nuisance structure rather than meaningful joint co-variation. Inspecting component loadings and visualizing principal movements therefore provides a straightforward way to detect such problems. Landmarks that consistently contribute little to variance, or that show patterns indicative of tracking noise, can be identified and removed before recomputing PCA, yielding a cleaner representation of movement.

Several studies highlight this diagnostic potential across different movement-science contexts. \citet{mejia2017pca} introduced PCA leverage and PCA robust distance as measures of outlyingness for detecting artifact-contaminated time points in high-dimensional signals, a principle directly applicable to identifying framewise irregularities in pose trajectories. \citet{baudet2014cross} demonstrated how PCA can reveal and correct cross-talk artifacts in gait kinematics, where marker placement errors cause spurious correlations between movement channels. \citet{van2022whole} compared PCA solutions derived from optical motion capture and inertial sensors, showing that discrepancies in higher-order components reflect system-specific noise characteristics and can be used to benchmark preprocessing quality. Together, these examples show that examining PCA structure is a practical method for identifying misalignment, noise sources, or other preprocessing inadequacies.

Taken together, PCA provides a flexible means of probing coordinated spatial structure, identifying dominant movement modes, and evaluating the quality of preprocessing steps. In practical terms, PCA is best used when the aim is to summarize shared spatial variance across many landmarks or to inspect the dominant organization of whole-body movement. When the primary interest is temporal organization, however, PCA should generally be treated with caution and is better applied to derived summary features---including RQA outputs---than to the raw trajectories themselves. It therefore occupies a distinct analytic role alongside amplitude-based metrics and nonlinear methods: whereas linear kinematic features quantify how much movement occurs, PCA reveals how that movement is distributed across the body, and recurrence analysis captures how it unfolds in time.

\subsection{Recurrence Quantification Analysis}\label{sec:rqa}

Linear metrics such as mean displacement, velocity, or variance capture the overall magnitude of movement but are insensitive to its temporal organization. Human movement signals are nonlinear, nonstationary, and shaped by coordination across multiple timescales \citep{bernstein1967coordination,newell1993variability,stergiou2011human,ducharme2018fractal,hausdorff2000fractal}. To characterize such structure, it is useful to adopt methods that examine not only \textit{how much} movement occurs but also \textit{how it is sequenced over time}. 

Recurrence analysis provides a general framework for detecting when a dynamical system revisits similar states \citep{zbilut_embeddings_1992}. Applied to pose data, it identifies moments when the configuration of landmarks (or derived features) returns to a previously observed pattern. The resulting recurrence plot—a two-dimensional matrix marking state re-encounters—makes it possible to visualize and quantify temporal structure that may not be visible in the raw signal \citep{eckmann_recurrence_1987}. Unlike purely spectral or correlational methods, recurrence analysis can reveal both regular cyclicity and irregular transitions, making it especially well-suited to movement data where stability and variability coexist \citep{riley1999recurrence,ramdani2013recurrence,blazkiewicz_recurrence_2023}.

\subsubsection{State-Space Reconstruction}

\begin{figure}
    \centering
    \includegraphics[width=\linewidth]{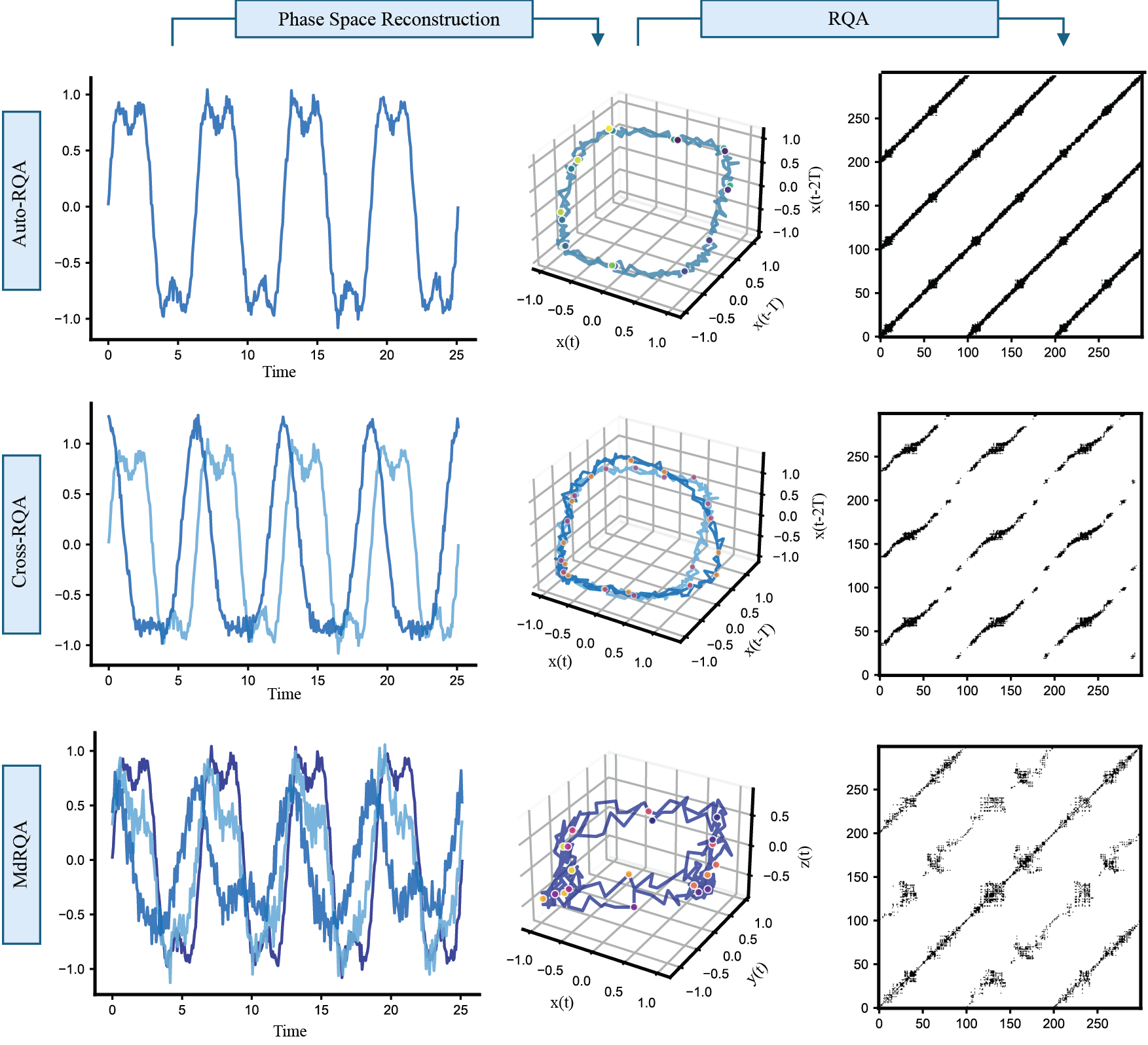}
    \caption{Illustration of recurrence quantification analysis (RQA) workflows. Each approach begins with a time series (left), which is reconstructed into a state space via time-delay embedding (middle), and then converted into a recurrence plot (right). Auto-RQA (top row) quantifies recurrent structure within a single trajectory, capturing the temporal organization of one system’s dynamics. Cross-RQA (middle row) embeds two trajectories into a shared phase space and identifies when one system revisits states previously visited by the other, allowing coordinated patterns to be quantified. Multidimensional RQA (bottom row) embeds multiple time series simultaneously, treating them as components of a single dynamical system to assess collective recurrence structure across signals. Together, these methods extend recurrence analysis from individual dynamics to coupled and high-dimensional systems, providing flexible tools for movement and coordination research.}
    \label{fig:rqa_state_space}
\end{figure}

To apply recurrence analysis, time series data must first be reconstructed into a state space that represents the system’s dynamics. This step, known as delay embedding, transforms a univariate signal into a higher-dimensional trajectory where recurrent states can be identified. The process is grounded in Takens' theorem \citep{takens1981}, which establishes that a time-delayed representation of a scalar signal can reconstruct the topology of the original dynamical system. In essence, the history of a single measurement contains information about the full system state: observing how a landmark's position evolves over several time lags reveals underlying coordination patterns even without simultaneous access to all degrees of freedom.

Delay embedding requires two parameters: the embedding delay or time lag ($\tau$) and the embedding dimension ($m$). Together, these define the geometry of the reconstructed state space and directly determine which patterns will be visible in recurrence analysis. In practice, both parameters must be selected in a way that is robust to noise, sampling variation, and natural movement variability.

\paragraph{Embedding Delay \texorpdfstring{$(\tau)$}{(tau)}} 
The delay specifies the temporal spacing between coordinates when constructing each state vector:
$$\mathbf{x}t = [x_t, x_{t-\tau}, \dots, x_{t-(m-1)\tau}].$$
If $\tau$ is too small, successive samples will be nearly identical due to temporal autocorrelation, providing redundant information. If $\tau$ is too large, samples become independent but may skip over relevant dynamics. An optimal delay balances these extremes.

Average Mutual Information (AMI) provides a principled estimate by quantifying the information shared between $x(t)$ and $x(t+\tau)$ \citep{fraser_independent_1986}. In practice, AMI is computed by discretizing the signal and calculating mutual information $I(\tau)$ for a range of delays. The resulting curve typically decays from perfect mutual information at $\tau=0$ toward lower values as independence increases. The first minimum or, for noisy signals, the first plateau identifies the delay at which sample provide maximal new information \citep{wallot_calculation_2018,nishimoto_implicit_2024}.

\begin{figure}
    \centering
    \includegraphics[width=0.8\linewidth]{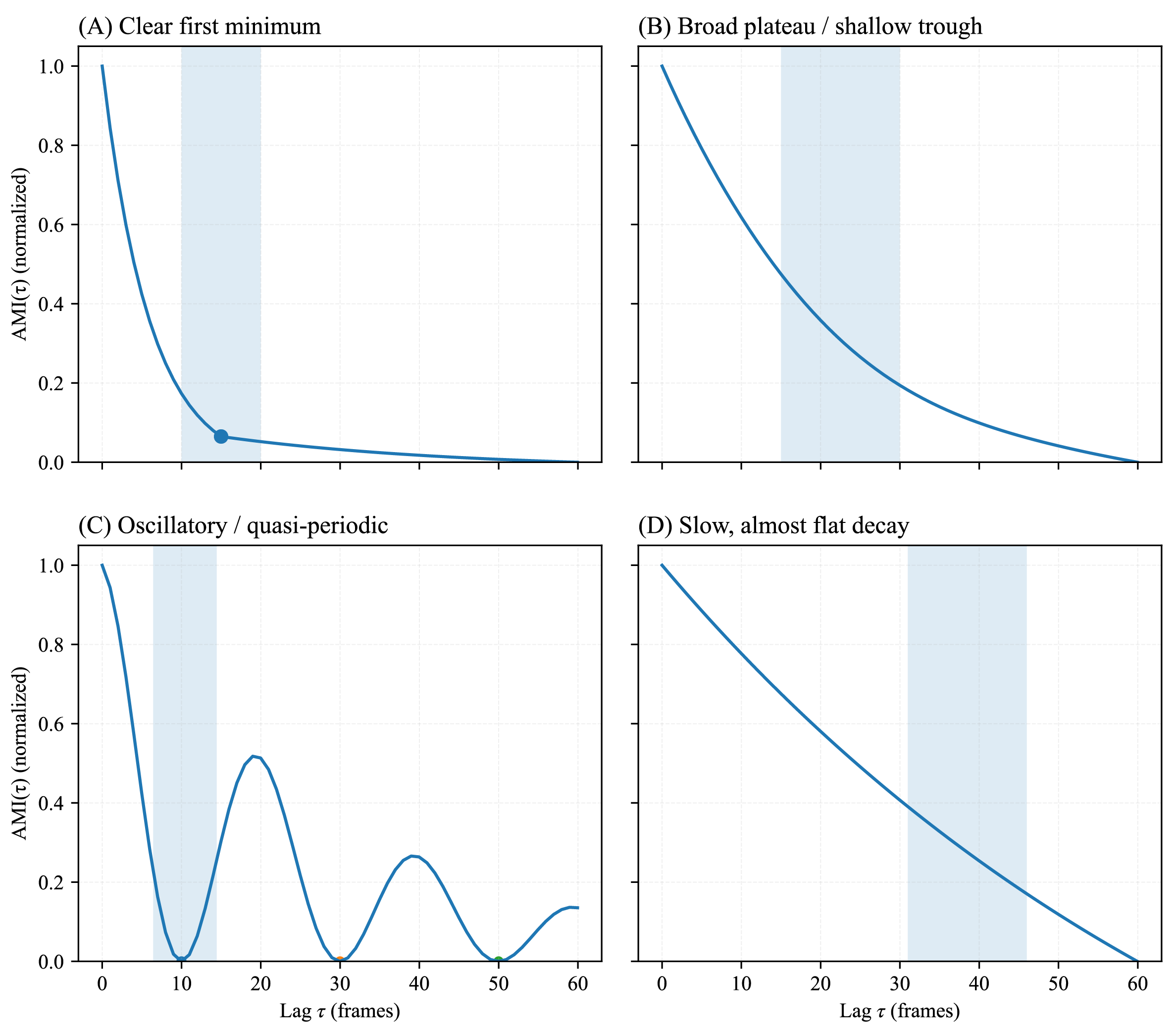}
    \caption{Illustrative Average Mutual Information (AMI) curves demonstrating common patterns encountered when selecting the embedding delay $\tau$ for pose-derived time series. (A) A clear first minimum, the canonical textbook case, where $\tau$ may be chosen directly at the first local minimum. (B) A broad plateau or shallow trough, characteristic of many natural movement signals; here a range of delays provides comparable information content, and robustness should be assessed across this interval. (C) An oscillatory or quasi-periodic AMI profile with multiple local minima; in such cases, the earliest stable minimum or plateau region is typically preferred. (D) A slow, almost flat decay indicating strong temporal redundancy at the sampling rate; delays may be selected at the onset of the plateau or based on a fixed AMI threshold. Across all panels, shaded bands denote plausible delay ranges, emphasizing that for non-rhythmic pose data, selecting a single $\tau$ is rarely justified. Robust analyses should verify that substantive results are stable across a small set of reasonable delays. The shaded bands are illustrative rather than prescriptive and are intended to denote plausible regions rather than precise optima.}
    \label{fig:ami_examples}
\end{figure}

Because pose-derived signals are often irregular and non-rhythmic, AMI curves frequently lack a sharp minimum and instead show broad or shallow troughs, oscillatory patterns, or slow decays (Fig.~\ref{fig:ami_examples}). For this reason, we recommend examining AMI across a representative sample of time series rather than relying on a single example. Modern software enables computing AMI profiles for dozens or hundreds of signals at negligible computational cost, allowing researchers to identify delays that are stable at the sample level. For pose data sampled at 30–60 Hz, embedding delays typically fall in the range $\tau = 10$–$30$ frames (166–1000 ms), varying with movement speed and signal smoothness; in the case studies below, $\tau$ ranged from 10 (Case 3) to 20 frames (Case 1). These considerations underscore that embedding delays should be selected from a plausible range and tested for robustness rather than fixed at a single value.

\paragraph{Embedding Dimension \texorpdfstring{$(m)$}{(m)}} 
The dimension determines how much temporal history is included in each reconstructed state. Too few dimensions fail to unfold the attractor, causing distant states to appear artificially close; too many introduce noise. The False Nearest Neighbors (FNN) method identifies the minimum dimension needed to avoid such degeneracies \citep{kennel1992}. FNN typically declines steeply at low $m$ and stabilizes once the attractor is adequately unfolded. As with AMI, $m$ should be estimated across a substantial portion of the dataset to ensure robustness \citep{bruijn_statistical_2009,wallot_calculation_2018}. For movement signals, $m$ values between 3 and 5 are common, and all three case studies below converged on $m=4$.

Given the heterogeneity and noise characteristics of pose data, selecting a single $\tau$, $m$, or recurrence radius is rarely sufficient. We therefore recommend evaluating results across a small grid of plausible parameter values and verifying that substantive conclusions are consistent across them. This practice helps ensure that findings reflect genuine movement structure rather than idiosyncrasies of parameter choice. In our case studies, we identified a narrow band of embedding delays and dimensions that produced stable recurrence metrics across the full dataset, and analyses were repeated across this range to confirm robustness.

In contemporary workflows, both AMI and FNN computations are fast enough to run at scale, removing the historical rationale for estimating parameters from only one or two representative signals. We view sample-level parameter estimation as a best-practice standard for RQA applied to high-dimensional pose data.

\subsubsection{Recurrence Detection}
% epsilon, distance rescaling, thelier window, min line length

Once the state space is reconstructed through delay embedding, the next step is to identify which states are recurrent–that is, which pairs of points in the trajectory are sufficiently similar to be considered ``revisits'' of the same configuration. This determination requires specifying a distance threshold and several safeguard parameters that prevent artifacts from dominating the recurrence structure. These choices shape the resulting recurrence plot and therefore the RQA metrics derived from it. 

\paragraph{Recurrence Threshold \texorpdfstring{$(\epsilon)$}{(epsilon)}}
Two states $\mathbf{x}_i$ and $\mathbf{x}_j$ are considered recurrent if their distance falls below a threshold $\epsilon$: $$
R_{i,j} =
\begin{cases}
1, & \text{if } \|\mathbf{x}_i - \mathbf{x}_j\| \leq \epsilon, \\
0, & \text{otherwise}.
\end{cases}
$$

The choice of $\epsilon$ is critical, as too large yields a saturated plot where nearly all points are recurrent, too small yields a sparse plot with insufficient structure to analyze. Two strategies exist for selecting $\epsilon$. 

The most common approach is to select a constant $\epsilon$ (e.g., 15-25\% of mean pairwise distance) across the sample and treat recurrence rate (RR) as a dependent variable. This method is straightforward, but it requires some tuning to identify an appropriate radius. A common heuristic is to target an RR of approximately 2–5\%.

Alternatively, one may adjust $\epsilon$ to achieve a target RR (again, typically 2-5\%), then use $\epsilon$ as a dependent variable. This approach ensures recurrence plots have comparable density across heterogeneous datasets and is especially useful when comparing movement scales that differ substantially in amplitude.

In the case studies below, we used fixed thresholds that varied across datasets and keypoint types. As illustrated, when working with many keypoints it can be useful to maintain a constant radius across the dataset, even if the resulting recurrence rates fall outside the typical range. Crucially, one should verify that observed effects and trends are not dependent on a narrowly tuned radius parameter. In practice, a fixed $\epsilon$ is often used as an initial choice, with recurrence rate serving as a diagnostic (typically falling in the range of 2--5\%). When differences in scale or amplitude make such comparisons unstable, adjusting $\epsilon$ to achieve a target recurrence rate may be more appropriate.

\paragraph{Distance Rescaling}
Before applying $\epsilon$, distances should typically be rescaled to make the threshold interpretable across different signals or conditions. Rescaling by mean distance divides all pairwise distances by their mean: $d'_{i,j} = d_{i,j} / \bar{d}$. With this rescaling, $\epsilon=0.20$ means ``20\% of the typical distance between states,'' providing an intuitive interpretation that generalizes across datasets. Alternatively, rescaling by maximum distance maps all values to [0,1] but can be sensitive to outliers. Applying $\epsilon$ directly to raw distances is only appropriate when absolute scale is theoretically meaningful. For pose data, where amplitude differences often reflect nuisance variance, mean rescaling provides a stable and interpretable reference for recurrence structure. As a practical default, we use mean rescaling unless differences in absolute scale are themselves of theoretical interest.

\paragraph{Theiler Window}
The Theiler window excludes a band around the main diagonal of the recurrence plot, removing matches between temporally adjacent states. This exclusion is only relevant in auto-RQA, because points close in time are almost always close in state space due to temporal continuity–not because the system has genuinely returned to a previous state. Without this safeguard, trivial autocorrelation inflates recurrence metrics artificially. 

The window width is typically set to $\tau$ (the embedding delay), ensuring that temporal neighbors excluded during embedding are also excluded from recurrence detection. A more conservative approach is the first zero-crossing of the autocorrelation function. For most pose signals, $\tau$ provides adequate protection. Visually, the Theiler window appears as an empty band centered on the diagonal, with width $2\tau+1$.

\paragraph{Minimum Line Length \texorpdfstring{$(l_{\min})$}{(l_min)}}
Diagonal and vertical lines in the recurrence plot reveal temporal structure: diagonals indicate predictable evolution, verticals indicate stable dwelling. However, very short lines often arise from noise or interpolation rather than genuine dynamical structure. The minimum line length parameter $l_{\min}$ specifies the shortest line to be counted.

Although the conventional starting point is to set $l_{\min}=2$ and count all possible diagonal lines—particularly in the context of physical systems—this choice can produce determinism values that are extremely high (95--100\%), obscuring meaningful variation. We therefore recommend using $l_{\min}=2$ as an initial default and then increasing it only if determinism is compressed near ceiling or shows insufficient variability within the analysis context.

\paragraph{Windowing and Temporal Resolution}
% why window, window length, window overlap, practical considerations
Most pose-derived signals are nonstationary: their structure changes over the course of a trial as participants adapt or shift strategy. For longer recordings (e.g., more than 1–2 minutes), applying RQA to the entire trajectory can obscure these changes by averaging over periods with distinct dynamical characteristics. For this reason, recurrence analysis is typically performed in sliding windows, each producing its own set of RQA measures and allowing temporal trends to be tracked.

The choice of window length reflects a trade-off. Longer windows yield more stable estimates of recurrence distributions—rules of thumb suggest a minimum of 1000 samples per window for reliable structure—yet they risk smoothing over meaningful fluctuations. Shorter windows improve temporal resolution but increase estimator variance. In practice, movement data sampled at 30–100 Hz often use windows of 30–120 seconds, though the optimal scale depends on the timescale of the behavior under study.

Window overlap is equally important. Without overlap, successive windows may begin and end at points that bear no dynamical relationship to one another, producing discontinuous or jagged RQA time series that reflect window boundaries rather than true behavioral change. Overlap ensures continuity by allowing adjacent windows to share a substantial portion of data, providing smoother temporal profiles and reducing sensitivity to arbitrary segmentation. A 50\% overlap is a common compromise: it doubles temporal resolution without inflating computational cost and preserves comparability across participants and conditions.

In practice, a useful default for pose-based RQA is to use sliding windows of sufficient length to retain at least \(\sim1000\) samples per window, with \textbf{50\% overlap} to ensure continuity between adjacent estimates. This provides a reasonable balance between metric stability and temporal sensitivity for many behavioral datasets. Shorter windows may be appropriate when the goal is to detect rapid reorganizations, whereas longer windows are preferable when the dynamics evolve more slowly or when recurrence estimates are unstable. In cases where behavior is expected to unfold across multiple timescales, researchers may also consider analyzing several window lengths within the same dataset to capture both fast and slow dynamics. Regardless of the exact values chosen, the same windowing scheme should be applied consistently across participants and conditions so that observed differences reflect genuine behavioral structure rather than artifacts of segmentation.

\subsubsection{RQA Metrics}

The recurrence plot provides a visual representation of temporal structure, but quantitative metrics are needed to characterize this structure objectively and enable statistical comparison. RQA metrics fall into two broad categories: density measures that quantify how often recurrence occurs, and structure measures that characterize how recurrent points are organized in time. Understanding how these metrics relate to one another and interpreting them collectively is essential for extracting meaningful dynamical information from movement data.

Recurrence Rate (RR) is the proportion of recurrent points, reflecting how often the system returns to similar configurations. Higher RR indicates more repetitive or constrained dynamics, though it provides no information about temporal organization. As \citet{marwan_recurrence_2007} notes, RR corresponds to the correlation sum in nonlinear time series analysis and effectively sets an upper bound for other measures, since subsequent metrics are computed as proportions of these recurrent points. Determinism (DET) quantifies the proportion of recurrent points forming diagonal lines of minimum length $l_{\min}$, typically set to two or three time steps. Diagonal lines indicate periods where trajectories evolve in parallel—not just starting from similar states but continuing to unfold similarly. DET therefore indexes predictability and temporal structure, with higher values indicating more deterministic dynamics \citep{webber_dynamical_1994}. 

Diagonal and vertical line structures capture different dynamical properties. Diagonal lines represent trajectory segments running parallel in phase space, quantifying predictability and system stability, whereas vertical lines represent segments remaining in the same phase space region, quantifying laminar phases and intermittent behavior \citep{marwan_recurrence_2007}. This means diagonal measures detect chaos-order transitions while vertical measures detect chaos-chaos transitions involving laminar phases. Average Diagonal Line Length (L) measures the mean duration of predictable sequences, with longer lines indicating sustained coordination or stable dynamical patterns. The maximum line length ($L_{\max}$) captures the longest sustained predictable episode in the window. Entropy (ENTR) characterizes the complexity of the diagonal line distribution. Higher entropy indicates diverse timescales—recurrence patterns of many different durations coexist—while lower entropy suggests dominance by a single characteristic timescale. ENTR complements DET since systems can have similar determinism but differ in whether structure reflects uniform repetition (low ENTR) or rich, multi-scale organization (high ENTR) \citep{charles2015recurrence}.

Laminarity (LAM) quantifies the proportion of recurrent points forming vertical lines, indicating that the system dwells in approximately the same state for consecutive time steps. Higher LAM reflects greater stability or persistence, which may indicate either adaptive control or maladaptive rigidity depending on context. Trapping Time (TT) is the average vertical line length, measuring how long the system remains in stable states. Research in postural control has established that LAM is often more sensitive to group differences than diagonal-line measures, as it captures motion fluidity during sway \citep{blazkiewicz_recurrence_2023}. A study of chronic ankle instability found that LAM was significantly lower in the affected population without corresponding changes in DET, indicating that the system was less likely to remain in specific states even though overall predictability appeared normal—a dissociation that would be missed by examining only one metric \citep{yen2022recurrence}.

These metrics are mathematically interdependent yet capture distinct aspects of system behavior, and they should therefore be interpreted collectively rather than in isolation. The combination of metrics reveals interpretable dynamical patterns that individual measures cannot distinguish. For example, high RR with low DET suggests stochastic recurrences resembling white noise, where many points recur but without sustained parallel trajectories, whereas low RR with high DET indicates rare but highly deterministic recurrences characteristic of structured periodic behavior \citep{marwan2011avoid}. The combination of high DET with high entropy signals a multi-stable regime with varied diagonal line lengths, indicating deterministic but complex dynamics, while high DET with low entropy signals stable, uniform dynamics such as periodic or quasi-periodic behavior \citep{charles2015recurrence}. \citet{richardson2007distinguishing} demonstrated experimentally that RR and $L_{\max}$ reflect independent aspects of coordination. Strengthening the attractor increased $L_{\max}$ with little change in RR, whereas adding noise increased RR without substantially altering $L_{\max}$. This independence shows why RQA is useful. Linear metrics like standard deviation cannot tell whether increased variability comes from weaker coupling or from added noise, but RQA distinguishes these sources by separating how often recurrences occur from how they unfold.

Although RQA metrics should be interpreted collectively, a small subset provides a useful entry point for most analyses. RR, DET, and either $L_{\max}$ or mean diagonal line length capture complementary aspects of recurrence density, predictability, and the duration of structured episodes, and therefore form a practical core set. Additional metrics can then be emphasized depending on the question: LAM and TT when persistence or intermittency is of interest, and ENTR when the diversity of timescales or complexity of structure is central. Importantly, prioritizing a subset does not imply that metrics should be interpreted in isolation, but rather that they serve as anchors for understanding the broader pattern of recurrence structure.

Clinical and applied studies consistently show that pathological movement patterns exhibit altered RQA profiles where multiple metrics change in interpretable combinations. Research on Parkinson's disease gait found that entropy, determinism, and average diagonal length all decreased with disease progression while divergence increased, a pattern interpreted as reflecting loss of quasi-periodicity and reduced adaptability \citep{afsar2018recurrence}. Studies of postural control in elderly adults report convergent reductions in both DET and ENTR in the anterior-posterior direction, indicating less predictable and less complex sway dynamics compared to young adults \citep{seigle2009dynamical}. Conversely, patients with persistent postural-perceptual dizziness exhibited higher DET and $L_{\max}$ than healthy controls despite no group differences in traditional linear measures, indicating a maladaptive rigid control strategy that RQA detected when amplitude-based metrics appeared normal \citep{kobel2023recurrence}. These examples suggest that the pattern across metrics reveals the nature of movement dynamics more fully than any single measure in isolation.

When RQA yields a large number of metrics across multiple features or conditions, dimensionality reduction techniques such as PCA can be applied to the resulting summary measures to identify common dimensions of variation. The choice of which metrics to emphasize depends on the research question, but interpreting them collectively—attending to their mathematical relationships and empirical dissociations—is essential for extracting meaningful insights into the temporal organization of movement. Mathematical definitions and computational details for all metrics are provided in the supplementary materials.

\subsubsection{RQA Variants}

\paragraph{Cross-Recurrence Quantification Analysis (CRQA)}
Auto-recurrence quantifies the temporal structure of a single signal, but many research questions focus on how two signals evolve together. Cross-recurrence quantification analysis (CRQA) extends RQA to pairs of time series by reconstructing each series into phase space and identifying when one trajectory visits states previously occupied by the other. Mathematically, auto-RQA evaluates $R(i,j) = \Theta(\varepsilon - \|x_i - x_j\|)$, which compares a time series to itself, while CRQA evaluates $CR(i,j) = \Theta(\varepsilon - \|x_i - y_j\|)$, which compares two different series. This change produces distinct structures. Auto-RQA is always symmetric and contains a filled main diagonal, whereas CRQA can be asymmetric and may lack a main diagonal altogether. These features reveal directional coupling and possible leader–follower relationships \citep{marwan_recurrence_2007}.

CRQA has been widely used to study coordination in movement, including gaze–head coupling, speech–gesture alignment, and interpersonal synchrony \citep{richardson2005looking,shockley2005cross,duong2024exploring,richardson2017nonlinear,romero2016using}. For example, \citet{richardson2005looking} showed that listener eye movements lag behind speaker eye movements by roughly two seconds during discourse, and this coupling predicted comprehension. \citet{shockley2005cross} found that conversational partners exhibit stronger postural cross-recurrence when speaking with each other than when speaking with a confederate, even though participants were unaware of this coordination. In these applications, CRQA metrics take the familiar form of auto-RQA metrics but now describe how two systems evolve in relation to one another.

A related method is joint recurrence analysis (JRA). In this approach, recurrence is computed separately for each signal and the resulting plots are multiplied element by element \citep{romano2004multivariate}. The joint recurrence matrix
\[
JR(i,j) = \Theta(\varepsilon_1 - \|x_i - x_j\|)\cdot \Theta(\varepsilon_2 - \|y_i - y_j\|)
\]
marks points where both systems are recurrent with themselves at the same moment. CRQA requires the two signals to occupy the same reconstructed space because it compares states directly. JRA does not require equal dimensionality and is therefore useful for combining heterogeneous data sources such as pose and physiological signals or for examining coordination among body regions with different measurement properties.

CRQA and JRA extend recurrence methods from single signals to coupled or multimodal systems, but they also introduce additional methodological considerations. Embedding parameters must be harmonized across signals. When delay or dimension estimates differ between series, it is generally safer to use the larger value because over-embedding is less problematic than under-embedding \citep{wallot_calculation_2018}. With appropriate parameter choices, both approaches provide effective tools for capturing coordination across multiple levels of behavior. In our case studies, CRQA identified decoupling between head and gaze movements in Case Study 1, and it characterized dyadic coordination in Case Studies 2 and 3.

\paragraph{Multidimensional Recurrence Quantification Analysis (MdRQA)}

CRQA reveals how pairs of signals evolve together, but many pose-derived behaviors involve coordination distributed across many features at once. Facial expression, for example, arises from the joint configuration of dozens of landmarks, and group coordination often reflects interactions among multiple individuals whose behaviors are interdependent. Multidimensional Recurrence Quantification Analysis (MdRQA) addresses this complexity by treating multiple time series as components of a single high-dimensional dynamical system \citep{wallot_multidimensional_2016}. 

The central insight behind MdRQA is that system-level coordination cannot always be inferred from all possible pairwise relationships. Suppose three pose features $x$, $y$, and $z$ each show moderate pairwise coupling. These dyadic links do not reveal whether the three features form a coherent collective pattern or whether each interacts independently with the others. MdRQA constructs a single trajectory in a multidimensional space where each feature contributes one or more dimensions. This approach captures global coordination patterns that transcend dyadic interactions. \citet{wallot_multidimensional_2016} demonstrated that group-level physiological coordination during collaborative tasks was detectable with MdRQA even when individual- and pairwise-level analyses showed no reliable effects. Group dynamics are not always reducible to pairwise components.

In MdRQA, the multidimensional state vector at time $t$ is
\[
\mathbf{W}_t = [y_{1,t}, y_{2,t}, \ldots, y_{N,t}],
\]
where each of the $N$ time series contributes its current value. Lagged coordinates can be added when embedding is applied. Recurrence is then computed as
\[
RP(i,j) = \Theta(\varepsilon - \|\mathbf{W}_i - \mathbf{W}_j\|),
\]
which identifies moments when the full multidimensional system returns to similar collective states. This differs fundamentally from concatenating embeddings or averaging over multiple CRQA results. MdRQA reveals when the system as a whole—not merely individual pairs of features—revisits comparable configurations (see Figure~\ref{fig:rqa_state_space}).

For pose data, such recurrences can reflect repeated collective expressions, multi-joint coordination patterns, or stable group movements. MdRQA has also been extended to Multidimensional Cross-Recurrence Quantification Analysis (MdCRQA), which compares two multivariate systems and identifies when they occupy similar regions of their respective phase spaces \citep{wallot2019multidimensional}. This extension has been used to quantify coordination between groups or between different modalities such as movement and physiology. Empirical comparisons show that CRQA applied to feature pairs and MdRQA applied to the full feature set can lead to different substantive conclusions about coordination strength and stability. MdRQA often detects system-level patterns that pairwise analyses miss \citep{wallot2018analyzing}.

These methods are therefore best chosen according to the structure of the systems under study: CRQA is most appropriate when examining coordination between two comparable signals, JRA when the systems are heterogeneous and cannot be meaningfully embedded in a shared state space, and MdRQA when the question concerns collective dynamics across multiple coupled signals. The latter, however, comes with increased computational cost and may require dimensionality reduction or feature selection in high-dimensional pose data.

\paragraph{Additional Resources and Methods}

Comprehensive computational tutorials on implementing RQA, CRQA, and MdRQA are available for readers who wish to explore these methods in greater depth \citep{macpherson2024advanced,wallot2017recurrence,wallot_multidimensional_2016,WallotLeonardi2023}. More detailed mathematical treatments of recurrence analysis are provided in \citet{marwan_recurrence_2007} and \citet{charles2015recurrence}. %All case studies presented here were implemented using an open-source Python toolbox we developed specifically for applying recurrence methods to pose data \citep{}.

A variety of nonlinear time-series techniques complement recurrence-based approaches. Detrended Fluctuation Analysis (DFA) quantifies long-range correlations and fractal scaling, revealing whether variability reflects random fluctuations or structured persistence across timescales \citep{peng_quantification_1995}. Sample Entropy and Multiscale Entropy quantify regularity and multi-scale complexity, offering information about predictability that is not captured by recurrence structure \citep{richman_physiologicaltimeseriesanalysis_2000,costa2002multiscale}. These measures provide valuable alternatives when specific aspects of dynamics are of interest.

Across auto-RQA, CRQA, and MdRQA, the choice of method depends on the research question and the structure of coordination under investigation. Auto-RQA reveals the internal temporal organization of individual signals, CRQA captures coordination between pairs of signals, and MdRQA extends this to collective dynamics across multiple signals. In practice, recurrence analysis relies on a sequence of interdependent decisions. Embedding parameters, distance metrics, thresholding schemes, and line criteria each shape the resulting measures. Small changes in these values can meaningfully alter recurrence plots, so transparent reporting and robustness checks are essential. When implemented with care, recurrence-based methods offer a flexible framework for quantifying the temporal structure of movement, revealing patterns of stability, variability, and coordination that are invisible to linear metrics alone.

With feature selection, preprocessing, and recurrence analysis established, we now turn to case studies that demonstrate how the proposed framework is instantiated in practice through a complete analysis pipeline. The following three studies illustrate different aspects of the framework and cover single-person 2D facial dynamics, dyadic 2D face and upper-body coordination, and 3D dyadic full-body coordination.

\section{Case Studies}
\begin{figure}
    \centering
    \includegraphics[width=\linewidth]{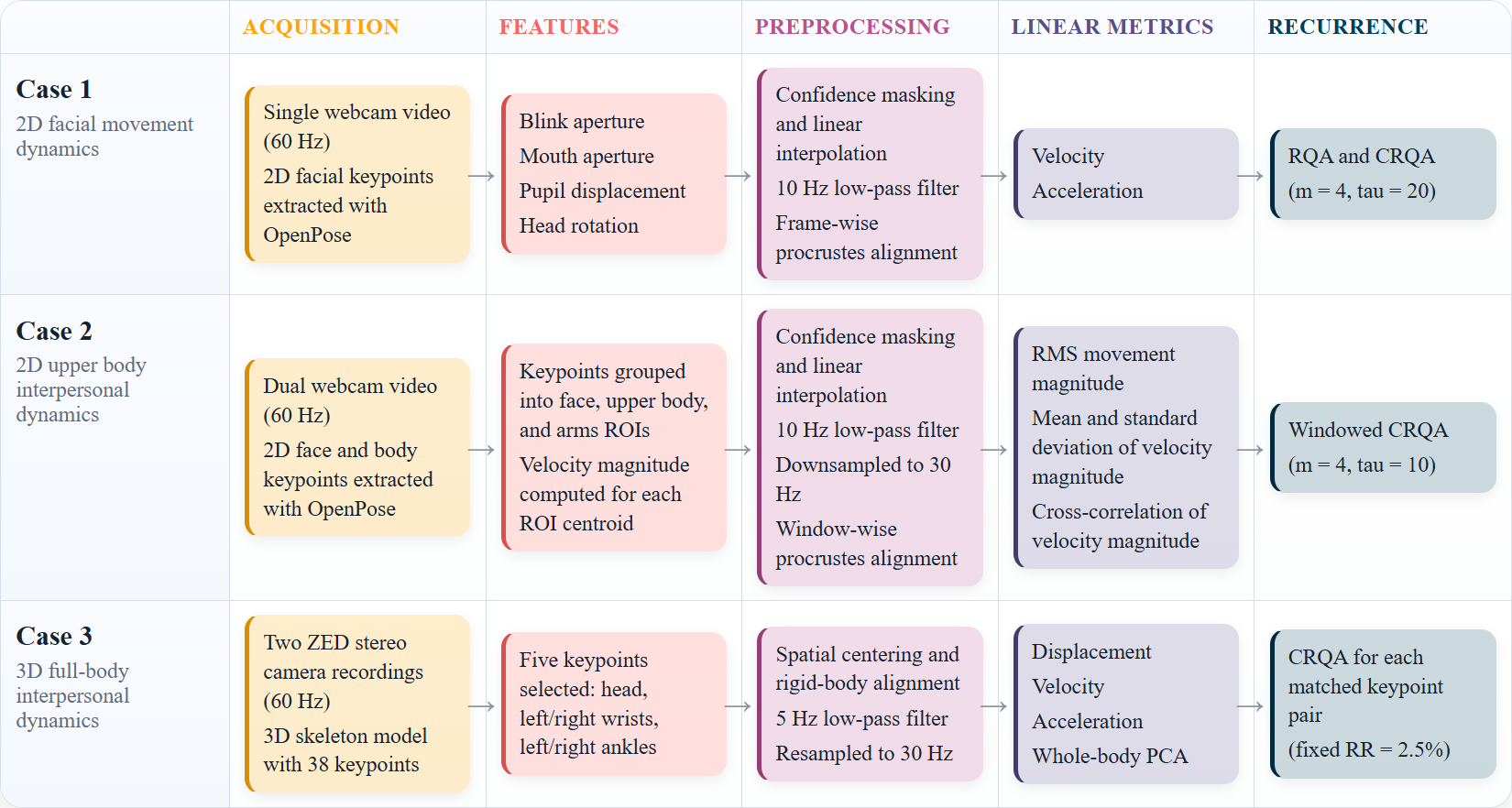}
    \caption{Analysis pipeline for pose-based behavioral research. The five-stage workflow is applied across three case studies to demonstrate adaptability across recording configurations and movement scales. Stages 1–3 represent the common preprocessing sequence, including low-pass filtering (LPF), confidence-based masking, and Procrustes alignment to ensure spatial and temporal standardization. Stage 4 extracts linear kinematic features to quantify movement magnitude and variability. Stage 5 employs recurrence quantification analysis (RQA/CRQA) to characterize the temporal organization and coordination dynamics of the behavioral signal.}
    \label{fig:analysis_pipeline}
\end{figure}

The following three case studies demonstrate the pipeline's adaptability across different pose estimation contexts and research questions. Case 1 examines single-person 2D facial dynamics during cognitive load manipulation, illustrating the accessibility of webcam-based pose estimation and the application of cross-recurrence analysis to intra-individual coordination (gaze-head coupling). Case 2 investigates dyadic 2D upper-body coordination during conversations in noisy environments, extending cross-recurrence methods to interpersonal coordination, and demonstrating techniques for handling multi-person pose data from separate camera views. Case 3 explores 3D dyadic full-body coordination in a mirror game task, showcasing stereo camera-based pose reconstruction and the application of multidimensional recurrence analysis to capture collective movement dynamics across different visual coupling conditions. An overview of the pipeline applied in each case study is illustrated in Figure~\ref{fig:analysis_pipeline}.

Each case study highlights different aspects of the pipeline: pose acquisition challenges (single webcam vs. dual webcams vs. stereo cameras), feature representation strategies (facial ROIs vs. upper-body keypoints vs. full-body 3D coordinates), and analytical approaches (auto-RQA vs. CRQA vs. MdRQA). Together, they illustrate how the same methodological framework can be adapted to diverse experimental contexts while maintaining analytical rigor and practicality.

All studies were conducted with approval from Macquarie University's Human Research Ethics Committee. For Cases 1 and 2, detailed experimental procedures and complete statistical results are reported in their respective publications \citep{sale2026facial,macpherson2026inprep}; the focus here is on pose estimation methodology and pipeline implementation. Case 3 presents a complete methodological description (see Appendix) as this work has not been published elsewhere. All preprocessing and recurrence quantification analyses across the three case studies were implemented using a custom Python package that is openly available (see Supplementary Materials). 

\subsection{Case 1: 2D Facial Pose}\label{sec:dataset1}

As a first demonstration of the analysis pipeline, we examined two-dimensional facial pose data collected during a multitasking environment designed to elicit varying cognitive load. The aim was to determine whether subtle changes in facial dynamics could be detected through standard markerless pose estimation using only a consumer webcam. This case study illustrates the accessibility of pose-based behavioral analysis and demonstrates how cross-recurrence methods can reveal coordination patterns within a single individual.

\begin{figure}
    \centering
    \includegraphics[width=\linewidth]{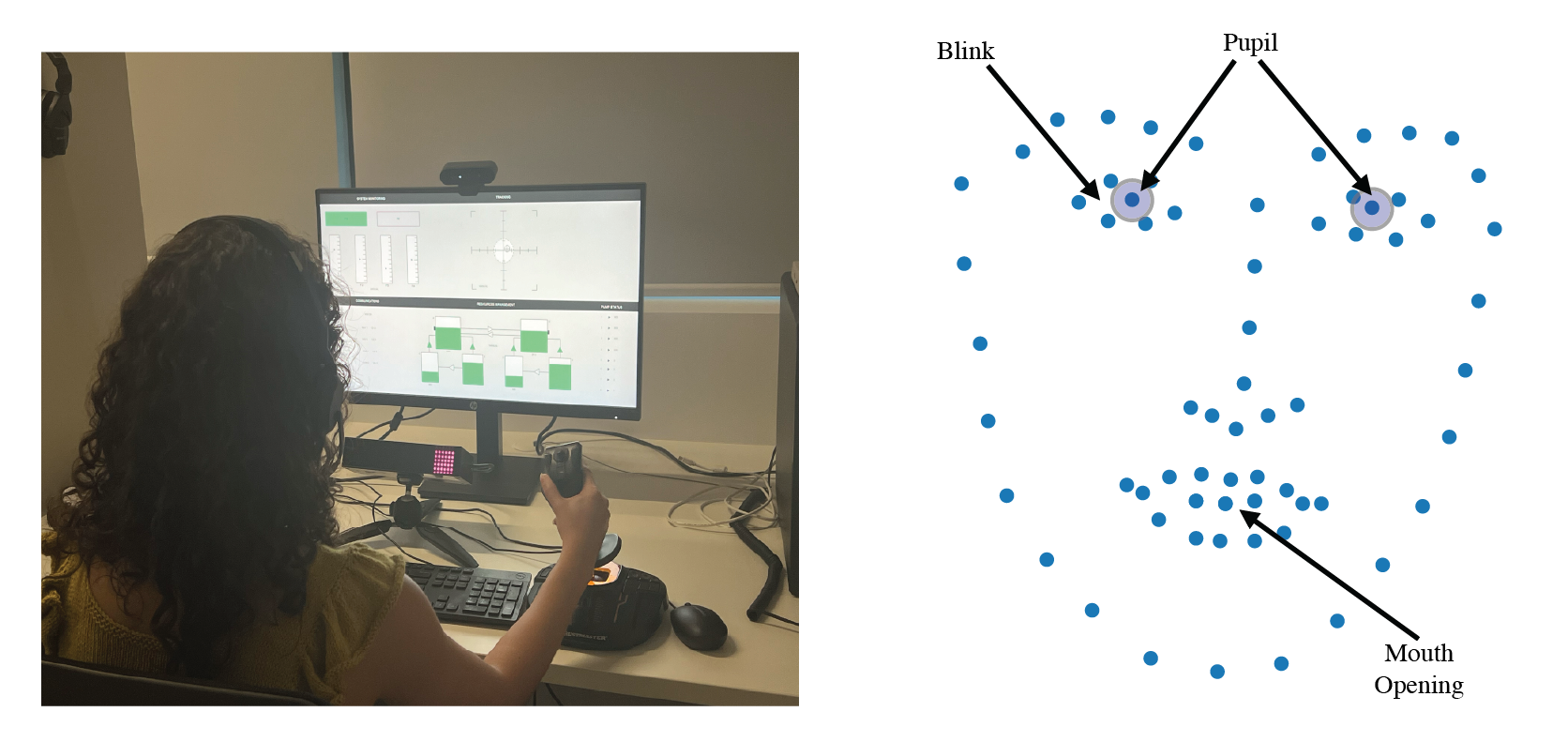}
    \caption{Experimental setup and facial pose features used for analysis. Left: Participant performing the Multi-Attribute Task Battery (MATB) aviation simulation while facial behavior was recorded with a consumer-grade webcam during multitasking under varying cognitive load. Right: Schematic illustration of the derived facial pose features used in analysis. Two-dimensional facial landmarks were grouped into functionally meaningful composites capturing blink aperture, mouth opening, pupil displacement, and rigid head motion (translation and rotation). These features formed the basis for subsequent linear kinematic analysis, recurrence quantification analysis (RQA), and cross-recurrence quantification analysis (CRQA) of gaze-head coordination.}
    \label{fig:matb_method}
\end{figure}

\subsubsection{Method}
Seventy-two participants completed the Multi-Attribute Task Battery (MATB) aviation simulation (OpenMATB v1.3.0; \citealp{cegarra_openmatb_2020}) while their faces were recorded at 60 Hz using a Logitech Brio webcam (Figure~\ref{fig:matb_method}). The MATB required simultaneous management of four subtasks: tracking a moving target with a joystick, monitoring system gauges, responding to radio communications, and managing fuel tank levels. Cognitive workload was manipulated across three eight-minute blocks (Low, Moderate, High) by varying task difficulty parameters.

\paragraph{Preprocessing}

Videos were processed offline using OpenPose's 70-keypoint facial landmark model \citep{cao2019openpose}, which estimates 2D pixel coordinates and confidence scores for anatomical landmarks spanning the eyes, eyebrows, nose, mouth, and face contour. This model was selected for its detailed facial coverage—enabling capture of subtle movements across multiple facial regions—as well as its ease of deployment and computational efficiency, making it suitable for scalable monitoring applications where real-time or near-real-time processing may be required.

Following the preprocessing procedures outlined in \ref{sec:generalmethod}, landmarks with confidence scores below 0.30 were masked as missing, and short gaps of 60 or fewer consecutive frames (1 second at 60 Hz) were filled using linear interpolation—as per the limits established by our missing data guidelines given the embedding delay of 20 frames. A zero-phase 4th-order Butterworth low-pass filter with a 10 Hz cutoff attenuated high-frequency noise, and coordinates were normalized to screen dimensions to yield resolution-independent values in the unit interval.

A particular challenge in this dataset was disentangling rigid head motion from facial expression. We addressed this through Procrustes superimposition, constructing a global reference template from four stable facial landmarks (nose tip [30], side of nose [31], top right eyelid [37], bottom left eyelid [46]) averaged across all frames and participants. Each frame's landmark configuration was aligned to this template, with the transformation parameters themselves—translation components, rotation angle, and scale factors—extracted as features quantifying head movement, while aligned landmark coordinates provided expression features with head motion factored out.

Rather than analyzing all 70 landmarks directly, we extracted anatomically motivated composite features representing functionally distinct facial regions, as described in the dimensionality reduction section of the General Method. Blink aperture was computed as the vertical distance between upper and lower eyelid landmarks, averaged across both eyes. Mouth aperture was measured as the Euclidean distance between upper and lower lip landmarks. Pupil displacement was calculated by determining eye centers from eye contour landmarks, then computing pupil landmark offsets from their respective centers, decomposed into x and y components with additional scalar magnitude. Additional features derived from Procrustes parameters included head translation magnitude, rotation angle, scale factors, and combined head motion magnitude.

\paragraph{Linear Kinematic Metrics}

For each feature, kinematic derivatives (velocity and acceleration) were computed using finite differences, expanding position measurements into dynamic representations. Features were aggregated into 60-second windows with 50\% overlap, computing nine summary statistics per window.

\paragraph{Recurrence Quantification Analysis}

Recurrence Quantification Analysis was applied to the unit-interval normalized time series following the procedures outlined in the General Method section on RQA. Embedding parameters were determined by running AMI and FNN on each feature, yielding $\tau = 20$ frames (333 ms) and $m = 4$, which were held constant across all features and participants. The recurrence threshold was set at 20\% of the mean pairwise distance ($\epsilon = 0.20\bar{d}$), a value that placed most features within the recommended 1–5\% recurrence-rate range. A few keypoint regions fell outside this range (such as blink and mouth movements), so we conducted additional radius-sensitivity analyses. For each affected feature we recalculated RQA metrics using feature-specific $\epsilon$ values that produced recurrence rates within 2–5\%. These checks showed that adjusting $\epsilon$ changed only the absolute magnitude of RQA measures and did not alter relative patterns, temporal trends, or statistical significance.

A Theiler window of two frames and a minimum diagonal line length of $l_{\min} = 4$ were selected based on pilot testing. Using the conventional $l_{\min} = 2$ produced determinism values in the 95–99\% range, which obscured meaningful variation in structure found at more conservative line length limits. Standard RQA measures including recurrence rate, determinism, laminarity, mean diagonal line length, and entropy were extracted from each recurrence matrix.

\paragraph{Cross Recurrence Quantification Analysis}

Cross-recurrence quantification analysis (CRQA) between pupil and mid-face X-axis signals assessed gaze-head coordination using shared embedding parameters ($m=4, \tau=20$ frames from cross-AMI averaged over 100 time series) with consistent threshold and line parameters from the auto-recurrence described above. 

\subsubsection{Results}
\begin{figure}
    \centering
    \includegraphics[width=0.7\linewidth]{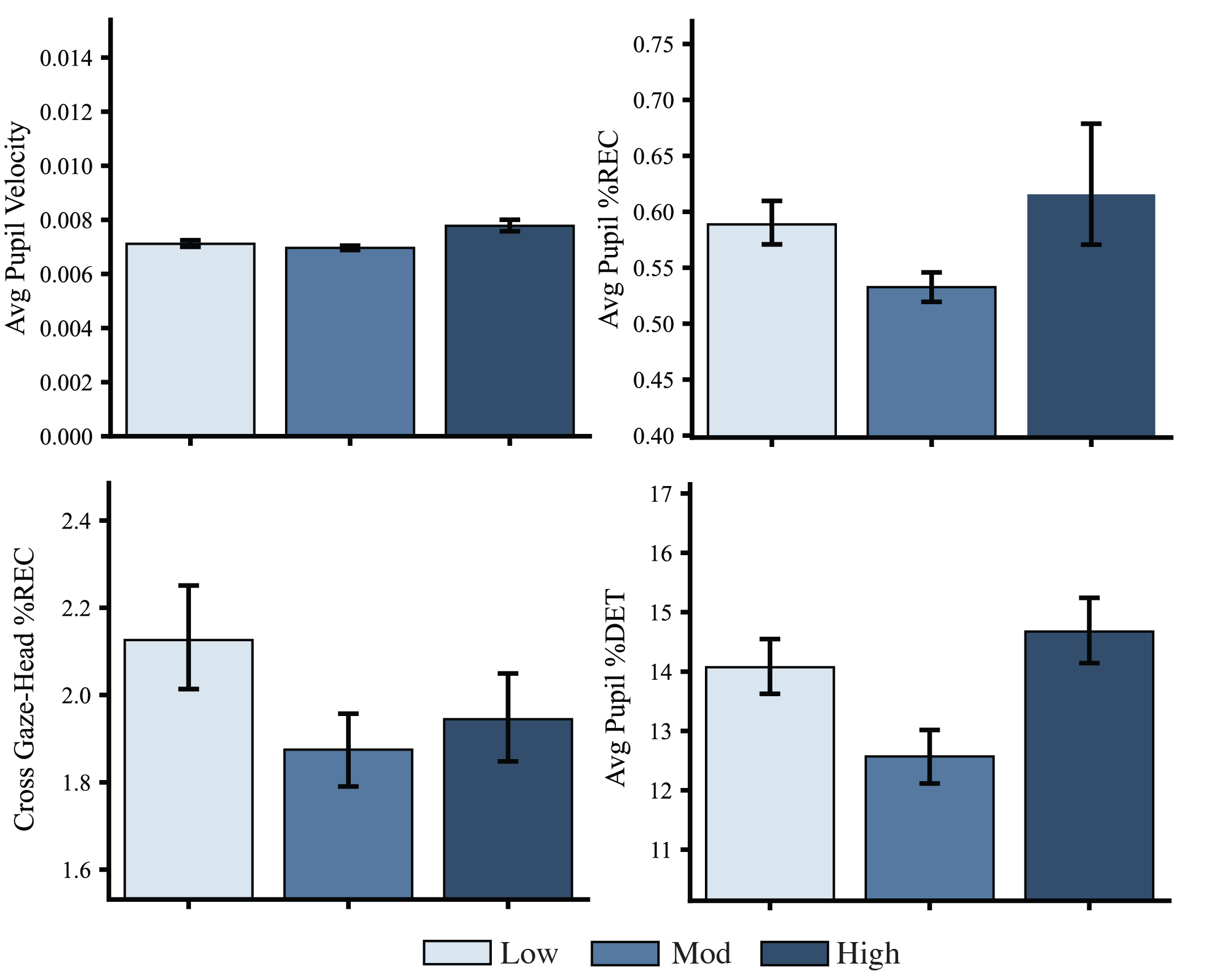}
    \caption{Representative linear and recurrence-based facial dynamics across cognitive load levels. Group-averaged metrics (mean $\pm$ SEM) illustrating load-dependent changes in facial behavior during the MATB task. From left to right: average pupil velocity, pupil recurrence rate (\%REC), cross-recurrence rate between gaze (pupil) and head motion, and pupil determinism (\%DET), shown for low, moderate, and high workload conditions. Linear kinematic measures show increased movement magnitude under high load. Recurrence-based measures reveal an initial reduction in repetitiveness from low to moderate load followed by structural reorganization at high load, reflected in increased determinism. In contrast, cross-recurrence between gaze and head decreases under load and does not exhibit comparable recovery, indicating progressive inter-system decoupling despite intra-system reorganization.}
    \label{fig:matb_results}
\end{figure}

Application of the analysis pipeline revealed systematic, load-dependent changes in facial dynamics at both linear and nonlinear levels (see Figure~\ref{fig:matb_results}). We present here the key findings that demonstrate the pipeline's capabilities; complete statistical results including all features, effect sizes, and model specifications are reported in the full study publication \citep{sale2026facial}. 

\paragraph{Linear Kinematic Metrics}

Analysis of linear kinematic features indicated systematic scaling with task demand. Across the experimental blocks, the velocity and acceleration of pupil and head movements increased significantly under higher task load, particularly in comparisons involving the high-load condition, demonstrating that greater task demand was associated with greater head and facial activity. Pupil velocity showed minimal change from low to moderate load but increased substantially at high load (RMS: H--L $d=0.33,p<.001$; H--M $d=0.39,p<.001$), with acceleration following the same pattern (RMS: H--L $d=0.31,p<.001$; H--M $d=0.37,p<.001$). Head rotation velocity increased primarily in comparisons involving high load (H--L $d=0.42,p<.001$; H--M $d=0.38,p<.001$), suggesting that high load demanded rapid switches in orientation. Mouth aperture increased progressively across both load transitions (velocity RMS: M--L $d=0.34,p<.001$, H--M $d=0.33,p<.001$), likely reflecting increased response rates to the communications task. Blink aperture decreased progressively (RMS: $d=-0.16$ from moderate to high, $p<.001$; mean: $d=-0.26$, $p<.001$), indicating more narrowed eye opening under load, while blink velocity and acceleration increased ($d=0.39$ and $0.34$ respectively from moderate to high, $p<.001$).

\paragraph{Recurrence Quantification Analysis}

Recurrence quantification analysis (RQA) revealed qualitative changes in the temporal structure of movement not discernible from linear kinematic measures alone. With increasing task demand, mid-face and pupil movements became more fragmented, evidenced by significant decreases in Recurrence Rate (RR) and Determinism (DET). 

From low to moderate load, pupil magnitude showed reduced recurrence ($d = -0.12, p = .021$) alongside substantial drops in determinism ($d = -0.28, p < .001$) and laminarity ($d = -0.31, p < .001$). However, from moderate to high load, determinism increased dramatically across most features despite recurrence remaining comparatively suppressed. Pupil magnitude determinism increased substantially ($d = 0.38, p < .001$), accompanied by sharp rises in entropy ($d = 0.34, p < .001$) and laminarity ($d = 0.40, p < .001$), indicating reorganization into more structured yet complex patterns. Head motion magnitude and rotation showed comparable reorganization (determinism increases: $d = 0.20$ and $0.19$ respectively, $p < .001$). Vertical pupil motion exhibited pronounced reorganization at high load (determinism: $d = 0.33, p < .001$; entropy: $d = 0.34, p < .001$), likely reflecting systematic downward scanning aligned with the vertical task layout, while horizontal pupil motion showed no comparable recovery and remained relatively fragmented.

This degradation of predictable movement patterns contrasted sharply with the dynamics of blinking. At high load levels, blink patterns became more structured, showing significant increases in both Determinism (DET: $d = 0.29, p < .001$) and Laminarity (LAM: $d = 0.26, p < .001$), alongside substantial increases in entropy ($d = 0.34, p < .001$) and mean line length ($d = 0.32, p < .001$). Mouth dynamics diverged from the typical two-phase pattern, showing consistent progressive increases in structure across both load transitions with no initial fragmentation phase. Recurrence increased steadily ($d = 0.16$ and $0.11, p < .024$), as did determinism ($d = 0.27$ and $0.23, p < .001$), likely reflecting the increasing frequency and regularity of verbal responses demanded by the communications task. Additional RQA measures and their complete statistical analyses are reported in the full publication \citep{sale2026facial}.

\paragraph{Cross Recurrence Quantification Analysis}

Furthermore, cross-recurrence analysis (CRQA) of gaze (pupil) and head (mid-face) movements revealed functional decoupling between the two systems, with cross-recurrence rate declining significantly under higher task demand. Coupling between gaze direction and head movement weakened from low to moderate load and remained reduced at high load. For the magnitude measure combining both axes, percentage recurrence declined from low to moderate load ($d = -0.14, p = .005$), indicating fewer returns to previously shared gaze-head states. Maximum line length showed reductions from low to moderate load ($d = -0.17, p < .001$) and from low to high load ($d = -0.11, p = .035$), while mean line length also declined from low to moderate load ($d = -0.15, p = .002$) before showing only modest change thereafter. Unlike individual facial features which showed compensatory reorganization at high load, cross-recurrence measures showed no comparably robust recovery. Determinism declined initially ($d = -0.13, p = .009$) with no significant change from moderate to high load ($d = 0.05, p = .450$), contrasting with the dramatic determinism increases seen in individual systems. Entropy decreased from low to moderate load ($d = -0.21, p < .001$) with only modest change thereafter ($d = 0.11, p = .026$ from moderate to high), far less robust than the entropy increases characterizing individual feature reorganization. The vertical axis showed particular interest: determinism exhibited a U-shaped trajectory—decreasing initially ($d = -0.16, p < .001$) then rebounding substantially at high load ($d = 0.16, p < .001$)—as did entropy and laminarity, suggesting late-stage reorganization in vertical-axis coordination despite overall coupling breakdown.

\subsubsection{Discussion}

The results of this case study demonstrate the utility of the proposed pipeline for extracting meaningful signatures from 2D facial pose data captured with minimal experimental overhead. The webcam-based setup required virtually no additional equipment beyond what participants already used for the computer task, illustrating the scalability of markerless pose estimation for naturalistic workload monitoring. In the context of a demanding multitasking environment, the pipeline successfully identified robust markers of cognitive load operating at multiple timescales. While not reported here when combined with machine learning approaches, these features were able to classify workload more reliably than task performance alone (see \citet{patil2026inprep,sale2026facial}). The combination of linear kinematics and nonlinear recurrence analysis provided a multi-layered account of behavior; while kinematic features confirmed that movement magnitude increased with load, RQA revealed a more nuanced reorganization of movement dynamics that would have been invisible to amplitude-based measures alone.

Specifically, the analysis highlighted RQA's sensitivity to complex, state-dependent changes in temporal organization. The method distinguished between two distinct and simultaneous responses to cognitive load: the fragmentation of facial and ocular movements and the concurrent reorganization of blink patterns into more structured rhythms. Most individual facial systems exhibited systematic adaptation under load, characterized by initial fragmentation followed by increased structural integration and complexity at peak demand. This finding reinforces the value of moving beyond simple statistical measures to characterize the qualitative structure of behavior, revealing that peak cognitive demands induce not continued degradation but rather emergence of new, more complex coordinative regimes.

Furthermore, this case study illustrates the power of cross-recurrence analysis for investigating coordination dynamics. While individual subsystems showed adaptive reorganization, the coordination between gaze and head movements told a different story: a progressive breakdown in coupling strength with only limited evidence of late-stage reorganization along the vertical axis. The observed degradation in gaze-head coupling under high load provides a quantitative marker of how normally integrated sensorimotor systems can decouple under task demands. This application highlights the pipeline's capacity to quantify the dynamic interplay between components within a single individual, demonstrating that the same analytic principle extends naturally to interpersonal contexts where cross-recurrence can measure coordination between individuals. The following case study explores this possibility by applying the pipeline to dyadic interaction data.

\subsection{Case 2: 2D Upper-Body Interpersonal Dynamics}\label{sec:dataset2}
For our second case study, we employed data from a larger project (see \citealp{luthy_communication_inprep, macpherson2026inprep}) designed to investigate the use of objective biomarkers to predict communication difficulty in background noise. For the present analyses, we examined two-dimensional facial and upper body pose data collected during unrestricted conversations in which participants were exposed to four levels of background noise (ranging from a quiet office environment to a loud party). The aim was to determine whether changes in background noise are reflected in the interpersonal dynamics of interactants, captured via cross-recurrence quantification analysis.

\subsubsection{Method}
Ninety-four participants (47 pairs; \textit{M}age = 23.11 years, \textit{SD} = 7.29, 26 man or male, 52 woman or female, 2 non-binary, 14 gender not reported), all with typical hearing, engaged in naturalistic conversation while exposed to four background noise scenes. Each pair was seated across from one another and fitted with open headphones that permitted near-transparent acoustic communication \citep{weisser_conversation_2019}. They were instructed to converse freely, with optional prompts provided on a sheet of paper. Participants received course credit or monetary compensation for their participation.

\begin{figure}[htbp!]
    \centering
    \includegraphics[width=0.7\linewidth]{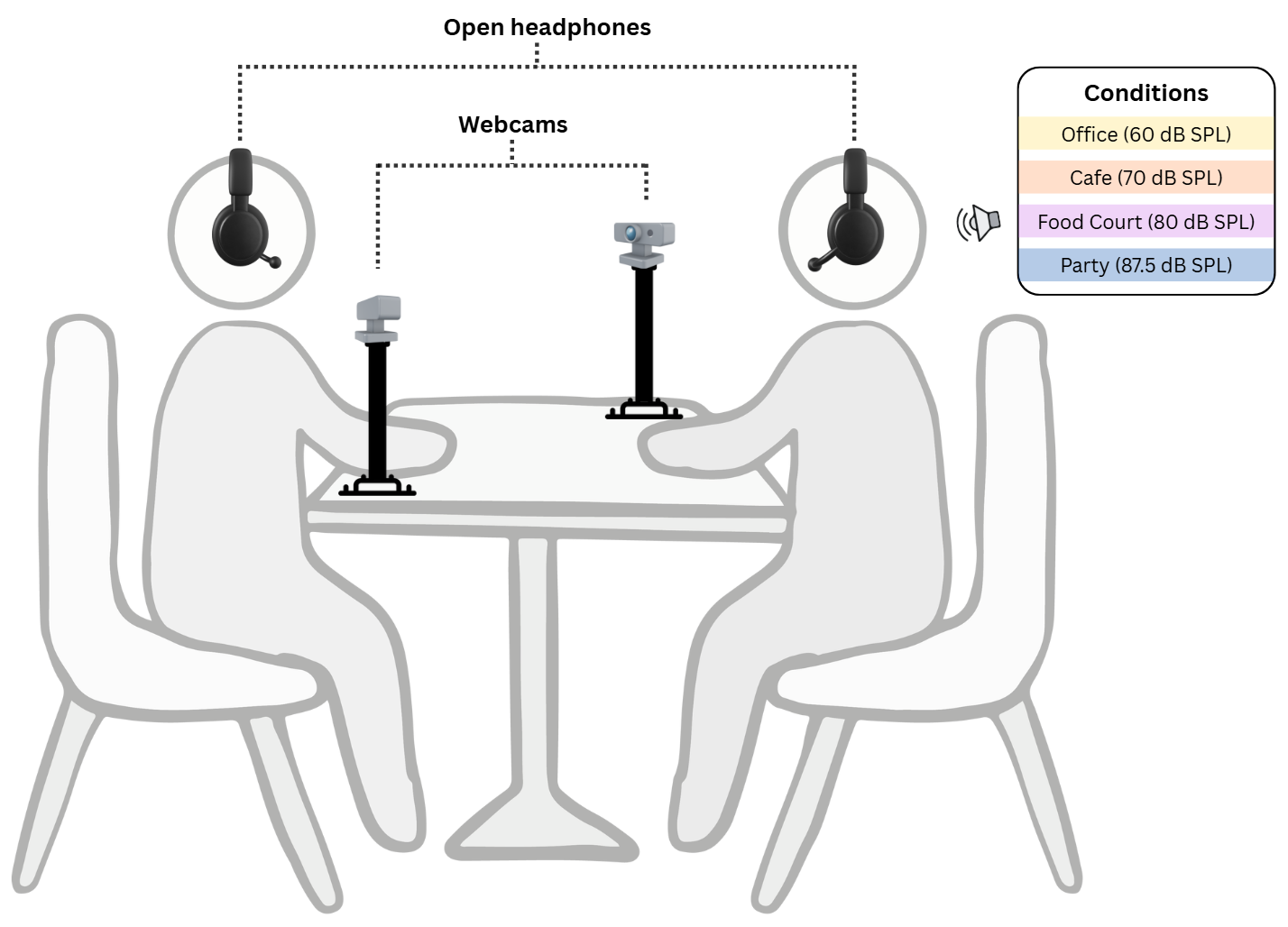}
    \caption{The experimental set-up for case study 2. Interactants sat facing one another and conversed while exposed to various background noise conditions: Office (60 dB SPL), Cafe (70 dB SPL), Food Court (80 dB SPL) and Party (87.5 dB SPL).}
    \label{fig:mosaic_method}
\end{figure}

The experimental design comprised four background noise conditions, resembling an office (60 dB SPL), café (70 dB SPL), food court (80 dB SPL), and party (87.5 dB SPL) (\citealp{weisser_arte_2019}; \citealp{weisser_conversation_2021}). These conditions were presented in a pseudo-random order, with the office condition appearing first (serving as a baseline) and between each of the other conditions (acting as a washout). The office trials lasted 5 minutes, while all other trials lasted 8 minutes (Figure~\ref{fig:mosaic_method}).

\paragraph{Preprocessing}

Movements were recorded at 60 Hz using two webcams, capturing each partner individually. Here, one partner was shot diagonally from the left, while their partner was shot diagonally from the right. The videos were processed offline using OpenPose’s 25-point body and 70-point face models to extract framewise 2D landmark coordinates. Keypoints prone to occlusion (outer face, lower body, hands) were excluded, leaving three anatomically defined regions of interest (ROIs): \textbf{centre-face} (body-model keypoints: 0, 15-16; face-model keypoints: 7, 9, 17-26, 31-69), \textbf{upper body} (body-model keypoints: 1-2, 5, 8) and \textbf{arms} (body-model keypoints: 2-7). 

Pixel coordinates of the selected keypoints were translated so that the video centre corresponded to [0,0], and then rescaled relative to the video resolution so that the frame edges mapped to [-1,1]. This process ensured a consistent origin and resolution across independent keypoint trajectories. Samples with low detection confidence were set to missing, and gaps of up to 1~s (60~frames at 60~Hz) were linearly interpolated. This limit corresponds to the conservative threshold $(m-1)\tau$ given the embedding parameters ($m = 4$, $\tau = 10$~frames at 30~Hz) implemented below. To attenuate high-frequency artifacts, a fourth-order Butterworth low-pass filter with a 10~Hz cutoff frequency (sampling rate = 60~Hz) was applied. Remaining missing values were temporarily interpolated prior to filtering to preserve continuity and then reinstated afterwards. Following filtering, time series were downsampled to 30~Hz to reduce computational load while preserving the temporal structure relevant for subsequent analyses.

To ensure comparability across camera angles, Procrustes alignment was then performed. Here, we constructed a single global template by averaging the position of each ROI keypoint across all valid frames in the dataset. Each trial was then segmented into overlapping 60~s windows (3,600~samples) with 50\% overlap. Within each window, coordinates were first centred on the mean nose position to remove gross positional bias. The mean pose of each window was then aligned to the global template via Procrustes transformation using singular value decomposition to estimate the optimal rotation, translation vector and uniform scaling factor. The resulting transformation was applied to all frames in the window.

Following alignment, movement dynamics were quantified at the level of the three ROIs defined above. Specifically, for each frame, keypoints belonging to the \textbf{centre-face}, \textbf{upper body}, and \textbf{arms} were spatially aggregated by computing the centroid (i.e., the mean x- and y-coordinates). Velocity (the first-order temporal derivative) was then computed from these centroid trajectories, and the magnitude of the resulting velocity vector was retained as a one-dimensional time series representing regional, sign-invariant movement intensity. This procedure reduced high-dimensional pose data to a set of interpretable ROI-level movement signals while preserving their temporal structure.

\paragraph{Linear Kinematic Metrics}

For each ROI, we extracted linear amplitude-based metrics from the magnitude time series on a window-by-window basis. These included root-mean-square (RMS) movement magnitude, as well as the mean and standard deviation of velocity magnitude, indexing overall movement energy and variability at the individual level.

To quantify linear coupling between interactants, we also computed a cross-correlation of velocity magnitude within each window. Here, partners' velocity-magnitude time series were z-scored within window, and the cross-correlation was evaluated over a bounded range of temporal lags. The cross-correlation value at lag 0 was then taken as an index of linear interpersonal coupling.

\paragraph{Cross Recurrence Quantification Analysis}

Interpersonal coordination was also quantified using cross-recurrence quantification analysis (CRQA). This was applied on a window-by-window basis to the z-scored velocity magnitude signals. Embedding parameters ($m = 4$, $\tau = 10$ frames) were selected using average mutual information (AMI) and false nearest neighbours (FNN) analyses and validated by examining the stability of the results across neighbouring parameter values ($m = 3-4$, $\tau = 10-20$). To maintain recurrence rates within the recommended range of approximately 1-5\%, a mean-rescaled radius of $20\%$ was used. The minimum diagonal line length was set to $l_{\min} = 2$. In contrast to Case Study 1, where larger line-length thresholds were required to avoid ceiling effects in determinism, pilot analyses indicated that $l_{\min} = 2$ yielded sufficient variability in determinism across windows in the present dataset.

Standard CRQA measures—including recurrence rate, determinism, laminarity, mean and maximum diagonal line length, entropy, and trapping time—were extracted for each window, yielding indices of interpersonal coordination across background noise conditions. 

\subsubsection{Results}

\begin{figure}[htp!]
    \centering
    \includegraphics[width=0.7\linewidth]{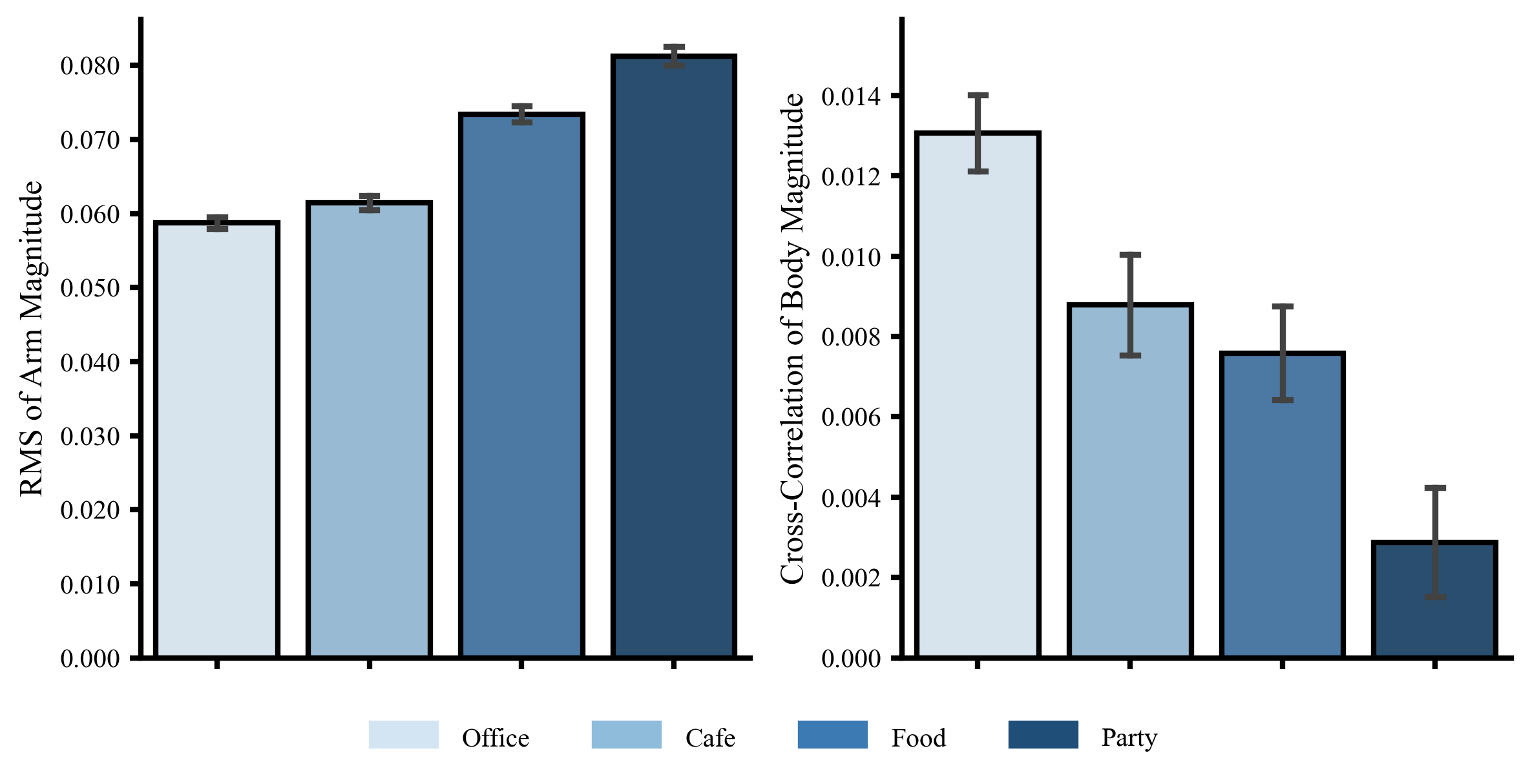}
    \caption{Select group-averaged linear metrics (mean $\pm$ SEM) illustrating changes in arm and upper body velocity magnitude across the background noise levels. From left to right: RMS of individual-level arm velocity magnitude and dyadic cross-correlation of upper body velocity magnitude. Individual linear kinematic measures generally increased as background noise level rose. However, the effects for dyadic linear measures were often inconsistent, with only the upper body showing a reduction in coupling as noise level increased.}
    \label{fig:linear_summary_mosaic}
\end{figure}

\paragraph{Individual and Dyadic Linear Metrics}

Linear mixed-effects models were fit to each linear metric with a fixed effect of background noise condition (Office as the reference). For individual-level metrics, models included random intercepts for pair and individual-within-pair. For the cross correlation, models included a random intercept for pair only. Across ROIs, movement intensity (RMS movement magnitude and mean velocity magnitude) and variability (SD velocity magnitude) increased with background noise. For RMS movement magnitude, Food Court and Party produced clear increases relative to the Office condition across the arms (Food Court $b = 0.014, SE = 0.002, t(451) = 7.07, p < .001$, Party $b = 0.024, SE = 0.002, t(451) = 11.69, p < .001$), upper body (Food Court $b = 0.006, SE = 0.001, t(451) = 7.25, p < .001$, Party $b = 0.012, SE = 0.001, t(451) = 13.13, p < .001$) and face (Food Court $b = 0.012, SE = 0.001, t(451) = 8.61, p < .001$, Party $b = 0.024, SE = 0.001, t(451) = 17.91, p < .001$). Moreover, both Food Court and Party always exceeded Cafe (all p < .001), and Party always exceeded Food Court (all p < .02). By contrast, no clear difference was uncovered between the two quietest conditions (Office and Cafe: all p > .15).

Mean and standard deviation of velocity magnitude showed similar effects, with all ROIs exhibiting increases in the Food Court and Party conditions relative to the Office and Cafe (all $p < .001$). The Cafe and Office conditions produced generally indistinguishable results, and where Cafe-related effects were present (e.g., mean velocity magnitude for face: $b = 0.024, SE = 0.012, t(451) = 2.06, p = .040$), these did not survive Tukey adjustment.

For dyadic linear coupling, no reliable effects were observed for either the arms or centre-face ROIs (all $p > .24$). For the upper body, interpersonal coordination of velocity magnitude decreased with increasing noise, with lower values in the Food Court ($b = -0.006, SE = 0.003, t(219) = -2.45, p < .015$) and Party ($b = -0.009, SE = 0.003, t(219) = -3.56, p < .001$) conditions relative to the Office. However, only Party remained significant following Tukey adjustment ($p = .003$). Together, these results indicate that increasing background noise systematically elevates individual movement velocity and variability, while simultaneous upper-body coupling shows a slight reduction under high noise levels (Figure~\ref{fig:linear_summary_mosaic}).

\paragraph{Cross Recurrence Quantification Analysis}

Linear mixed-effects models were also fit to each cross-recurrence (CRQA) metric with a fixed effect of background noise condition (Office as the reference) and random intercepts for pair. For the arms and upper body ROIs, CRQA metrics generally increased with background noise, particularly in the Food Court and Party conditions relative to the Office condition. For example, recurrence rate increased for both the arms (Food Court: $b = 0.735$, $SE = 0.280$, $t(221) = 2.623$, $p = .009$; Party: $b = 0.978$, $SE = 0.282$, $t(221) = 3.463$, $p < .001$) and upper body (Food Court: $b = 0.239$, $SE = 0.043$, $t(220) = 5.547$, $p < .001$; Party: $b = 0.285$, $SE = 0.043$, $t(220) = 6.568$, $p < .001$), with similar effects found for \%DET, maximum diagonal line length (Lmax) and entropy (Tukey-adjusted $p \le .003$).

\begin{figure}
    \centering
    \includegraphics[width=0.7\linewidth]{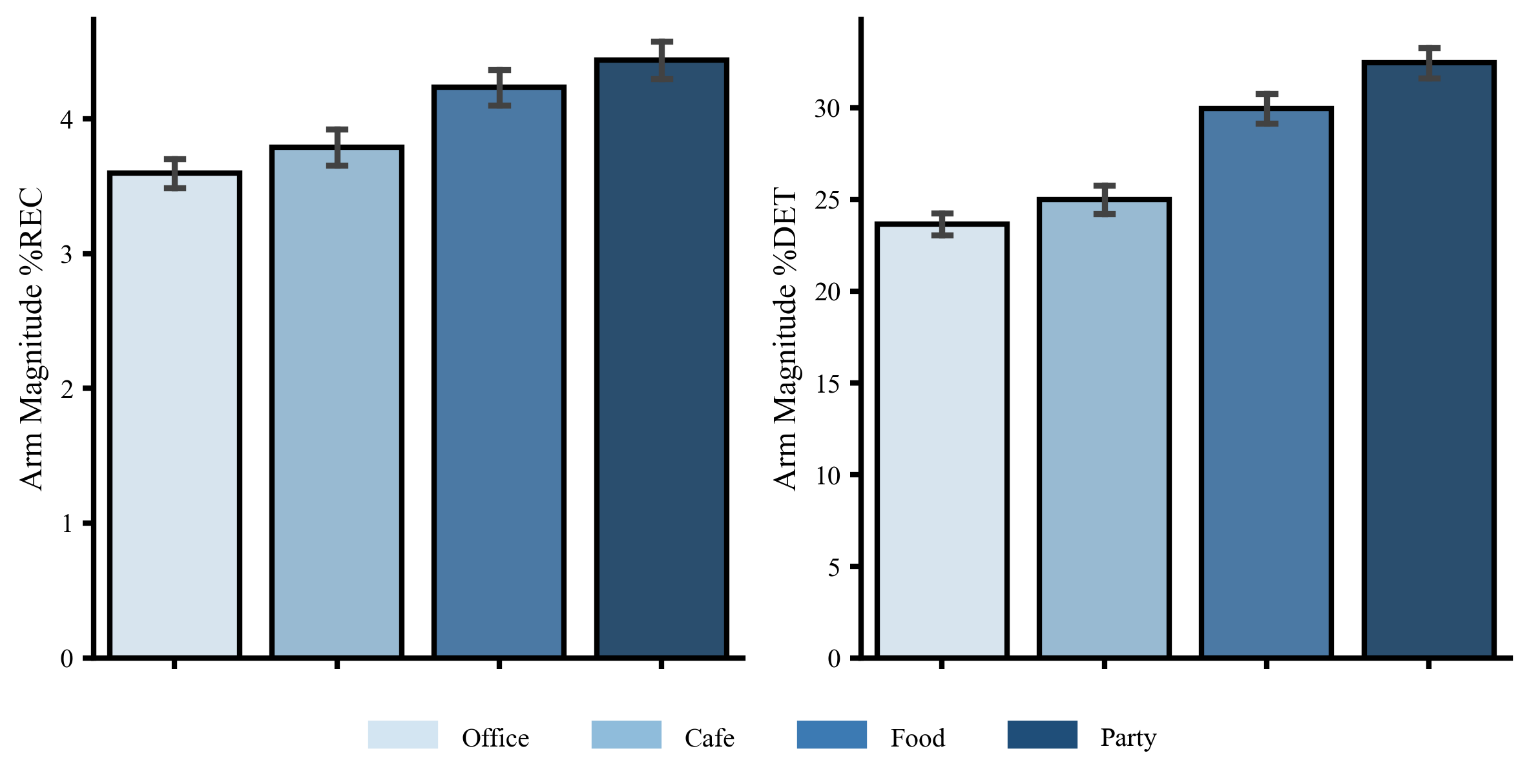}
    \caption{Select group-averaged CRQA metrics (mean $\pm$ SEM) illustrating changes in arm velocity magnitude across the background noise levels. From left to right: \%REC of arm velocity magnitude and  \%DET of arm velocity magnitude. Non-linear interpersonal coordination of the arms and upper body generally increased as background noise level rose.}
    \label{fig:rqa_summary_mosaic}
\end{figure}

By contrast, the face did not show reliable condition effects for recurrence rate (all Tukey-adjusted $p > .75$), though did exhibit increases in \%DET, Lmax, and entropy in the Food Court and Party conditions relative to Office (e.g., \%DET: Food Court $b = 4.730$, $SE = 1.398$, $t(221) = 3.383$, $p < .001$; Party $b = 7.520$, $SE = 1.408$, $t(220) = 5.340$, $p < .001$). These results indicate that increasing background noise enhances non-linear interpersonal coupling across the face, upper body, and arms, an effect that contrasts the linear coupling findings (Figure~\ref{fig:rqa_summary_mosaic}).

\subsubsection{Discussion}

This case study demonstrates that the proposed pipeline can be extended beyond single-participant analyses (case study 1), to capture meaningful patterns of interpersonal coordination from markerless facial, upper-body, and arm pose data during naturalistic conversation. By combining pose estimation with linear and non-linear time series analyses, we uncovered systematic changes in both individual movement dynamics and dyadic coordination as a function of increasing background noise.

Increases in background noise were associated with elevated movement intensity and variability at the individual level, particularly in the arms and upper body. In parallel, non-linear coordination metrics derived from CRQA revealed robust increases in recurrent structure, determinism, stability, and system complexity under higher noise conditions, replicating previous findings that interpersonal coordination becomes more stable (indexed via Lmax) in acoustically challenging environments \citep{miles2023behavioral}. Importantly, the present results show that these effects can be uncovered using markerless pose estimation in addition to magnetic motion tracking systems, increasing the flexibility and ecological validity of the design. This methodological shift is particularly valuable for applied and clinical contexts in which wearable sensors may be impractical or disruptive.

Notably, the combined use of linear and non-linear metrics provided complementary insights into the interpersonal dynamics of interactants that would not be apparent from either approach alone. Linear amplitude-based metrics captured robust increases in individual movement magnitude and variability with noise, while linear cross-correlation revealed modest reductions in upper-body coupling at the highest noise levels. In contrast, CRQA uncovered an \textit{increase} in non-linear coordination across the arms, upper body, and face. Together, these patterns suggest that while overt synchronous alignment between interactants may weaken under high noise, their movements simultaneously become more tightly organised, consistent with systemic adaptation to challenging environments.

Finally, the ROI-specific results highlight the utility of examining specific body regions when working with high-dimensional pose data. While the arms and upper body exhibited increased recurrence under noisy conditions, the face primarily showed changes in determinism, stability, and complexity rather than overall recurrence. This suggests that different anatomical regions may support distinct coordination strategies under acoustic challenge. Such distinctions would be difficult to detect without markerless pose estimation, dimensionality reduction, and region-specific aggregation, further underscoring the utility of the analytical choices implemented in the proposed pipeline. 

Overall, this case study illustrates how markerless pose estimation, when combined with principled preprocessing, alignment, and recurrence-based analyses, can be used to extract interpretable and theoretically meaningful indices of interpersonal coordination from naturalistic interaction data.

\subsection{Case 3: 3D Full Body Dyadic (Mirror Game)}\label{sec:dataset3}
The first two case studies utilized two-dimensional pose data acquired from standard RGB cameras. We then sought to scale the pipeline to three dimensions by applying it to full-body kinematics recorded during a dyadic coordination task (Mirror Game). In this task, pairs of participants engaged in short movement trials in which one partner (the leader) generated spontaneous upper-body motions while the other (the follower) attempted to mirror them as closely as possible. 

\subsubsection{Method}\label{sec:mirror_method}

\begin{figure}[htbp!]
    \centering
    \includegraphics[width=\linewidth]{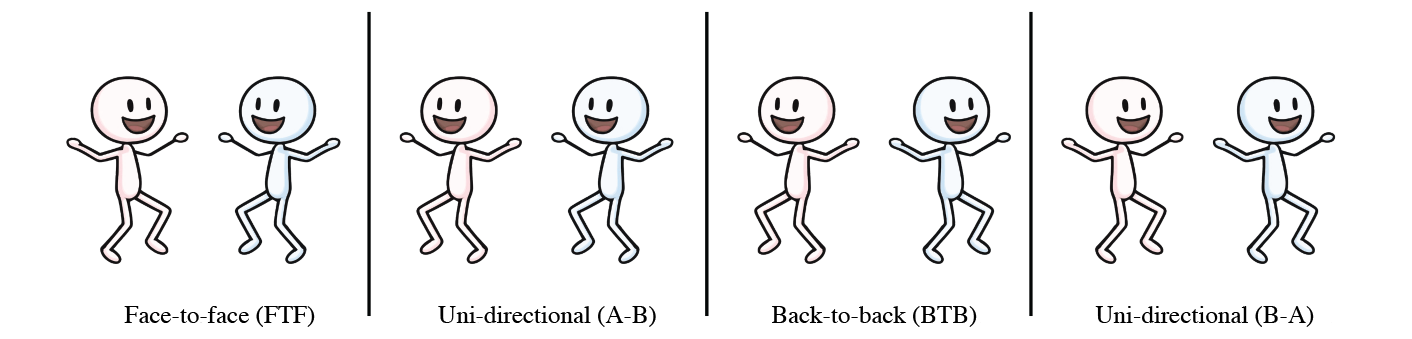}
    \caption{The three experimental conditions for the mirror game: face-to-face (FTF; mutual visual access), unidirectional (uni; asymmetric visual coupling with leader designation A or B), and back-to-back (BTB; no visual information).}
    \label{fig:mirror_game_task}
\end{figure}

A total of eighteen dyads completed the experiment. Visual coupling was systematically varied across trials by changing the participants’ orientation: \textit{back-to-back} (no visual information), \textit{unidirectional} (follower could see leader, leader faced away), and \textit{face-to-face} (full mutual visual access) (Figure~\ref{fig:mirror_game_task}). The experiment consisted of two blocks, one with each participant assigned as leader, and all trials lasted 30s. Movements were recorded at 60~Hz using two synchronized ZED stereo cameras \citep{stereolabs_zed_camera}, enabling full-body 3D pose estimation for both participants.

\paragraph{Preprocessing}
Raw 3D pose data were extracted using the ZED 38-keypoint full-body model, which provides 3D coordinates for major joints and body landmarks. The ZED recordings contained complete trajectories with no dropped frames or missing keypoints, so no interpolation or gap-filling was required.

All trajectories were first resampled to 30 Hz using cubic interpolation to match the temporal resolution used for other datasets in the project. Because leader and follower recordings occasionally differed in exact start/end times, resampling was followed by dyad-level temporal alignment: each trial’s two motion streams were trimmed to their overlapping time window, ensuring strict frame-to-frame correspondence.

To remove global translation and isolate movement structure, each frame was centered by subtracting the mean 3D position of the skeleton. Centering preserved relative geometry and dynamics while eliminating absolute location in space.

Next, a canonical body template was then constructed from the aggregated set of centered poses across all trials and participants. Each pose sequence was rigidly aligned to this template using Procrustes alignment with rotation allowed but no scaling. This step standardized body orientation across participants while preserving individual morphology and the intrinsic temporal evolution of each trial. 

After alignment, each keypoint trajectory was smoothed with a zero-phase 4th-order Butterworth low-pass filter with a 5 Hz cutoff, applied separately to x, y, and z coordinates for all 38 keypoints. Filtering at this stage removed frame-to-frame jitter while respecting the consistent spatial structure imposed by the Procrustes step.

The aligned-and-filtered 38-keypoint sequences were saved for downstream analyses. Although subsequent CRQA/MdRQA analyses use a reduced set of five keypoints (head, left/right wrists, left/right ankles), all trials underwent full-body preprocessing to support the global PCA described below. 

\paragraph{Principal Component Analysis}

To characterize dominant movement patterns across the full body, an exploratory Principal Component Analysis (PCA) was performed on the centered, Procrustes-aligned, and Butterworth-filtered pose data. All trials from all participants were concatenated into a matrix of size $(T_{\text{total},} \space 38\times3),$ where each row represented a single frame of a standardized 3D body configuration. This global PCA decomposed this high-dimensional space into orthogonal axes of spatial covariation–Principal Movements (PMs)–capturing shared low-dimensional structure in whole-body coordination across the dataset. The first 14 components explained roughly 96\% of total variance. 

Each trial was then projected into the PC space to visualize how strongly each PM was expressed over time. This step was used as a diagnostic to confirm data quality and reveal dominant movement modes. Importantly, PCA was used only for exploratory visualization. All nonlinear recurrence analyses were conduced on the predefined subset of five anatomically informative keypoints rather than on PCA scores.

\paragraph{Linear Kinematic Features}

Before, computing recurrence analyses, linear kinematic features were extracted to quantify how visual coupling influenced overall movement magnitude. From each preprocessed trial, the 3D trajectories of a subset of five anatomically informative keypoints (head, left/right wrists, left/right ankles) were used to compute frame-to-frame displacement, velocity, and acceleration. Displacement was defined as the Euclidean distance between successive frames, velocity as its first temporal derivative, and acceleration as the derivative of velocity. For each keypoint and each kinematic signal, both mean-based statistics (mean, standard deviation, maximum) and root-mean-square (RMS) values were computed to capture average magnitude as well as overall energetic contribution. In addition to per-keypoint features, global ``subset'' metrics were obtained by averaging across the five keypoints, providing a compact summary of whole-body movement amplitude. 

\paragraph{Cross Recurrence Quantification Analysis}

To quantify coordination structure between leader and follower, Cross-Recurrence Quantification Analysis (CRQA) was applied to five anatomically informative keypoints (head, left/right wrists, left/right ankles). For each keypoint, the 3D trajectory was collapsed into a scalar magnitude time series by computing the Euclidean norm across x, y, and z coordinates at each frame.

Before constructing recurrence matrices, delay embedding parameters were estimated independently for each trial. Average Mutual Information (AMI) was used to identify the first local minimum of mutual information, and False Nearest Neighbors (FNN) determined the minimum embedding dimension. Parameter estimates were computed separately for each keypoint's magnitude time series and then averaged across all trials, resulting in an embedding dimension of $m=4$ and time delay of $\tau=20$ frames. These parameters were applied to all subsequent recurrence analyses.

For each trial and keypoint, leader and follower magnitude time series were delay-embedded and z-scored before computing cross-recurrence matrices using Euclidean distance. The recurrence threshold was set to achieve a fixed recurrence rate of 2.5\% for each matrix. Recurrence quantification metrics were extracted using a minimum line length of 2 frames, computed separately for each of the five keypoints.

\subsubsection{Results}

\begin{figure}
    \centering
    \includegraphics[width=\linewidth]{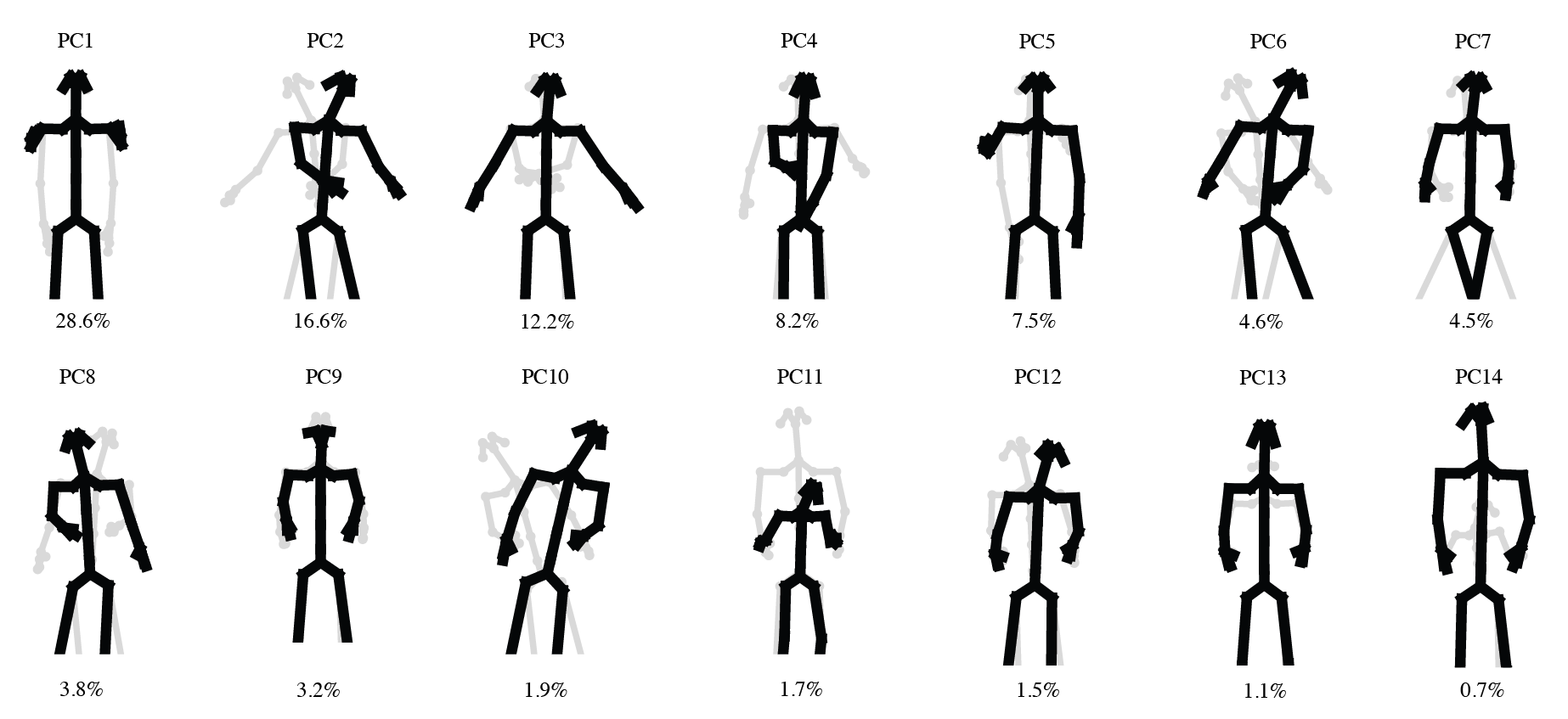}
    \caption{Principal Movements (PMs) extracted from Mirror Game full-body coordination. The first 14 principal components visualized as min (gray) vs. max (black) postures in frontal view. Each PM is scaled by an amplification factor to achieve a target RMS displacement of 0.25 units for visual clarity. The 14 components together account for the majority of postural variance across all dyads and conditions.}
    \label{fig:mirror_game}
\end{figure}

\paragraph{Principal Movements and Kinematic Metrics}\label{sec:mg_pca}

Following preprocessing, whole-body structure was examined using Principal Component Analysis (PCA) applied to the full 38-keypoint 3D trajectories. PCA revealed a low-dimensional postural space in which the first 14 principal movements (PMs) captured approximately 96\% of total variance (Figure~\ref{fig:mirror_game}). These PMs reflected coherent modes of multijoint covariation (e.g., sagittal arm motions, axial trunk rotations), confirming that body alignment and filtering preserved meaningful global structure.

To quantify how visual coupling influenced overall movement magnitude, linear kinematic features were extracted from a subset of five anatomically informative keypoints (head, left/right wrists, left/right ankles). For each trial, displacement, velocity, and acceleration time series were computed and summarized using both mean and root-mean-square (RMS) measures.

\begin{table}[htbp]
\centering
\caption{Condition effects on subset-level kinematic features from linear mixed-effects models}
\label{tab:mirror_game_subset_lme}
\begin{tabular}{llcccc}
\hline
Movement Type & Feature & Uni $\beta$ ($SE$) & $p$ & FTF $\beta$ ($SE$) & $p$ \\
\hline
\multirow{2}{*}{Displacement} 
 & Mean & 0.0022 (0.0005) & $< .001$ & 0.0031 (0.0005) & $< .001$ \\
 & RMS  & 0.0037 (0.0007) & $< .001$ & 0.0047 (0.0007) & $< .001$ \\

\multirow{2}{*}{Velocity} 
 & Mean & 0.067 (0.015) & $< .001$ & 0.093 (0.015) & $< .001$ \\
 & RMS  & 0.110 (0.021) & $< .001$ & 0.141 (0.021) & $< .001$ \\

\multirow{2}{*}{Acceleration} 
 & Mean & 0.740 (0.162) & $< .001$ & 0.889 (0.162) & $< .001$ \\
 & RMS  & 1.237 (0.252) & $< .001$ & 1.380 (0.252) & $< .001$ \\

\hline
\end{tabular}
\begin{flushleft}
\footnotesize Note. Back-to-back was the reference condition. Subset features were computed from the predefined five-keypoint subset (head, left/right wrists, left/right ankles) and averaged across keypoints prior to model fitting. Uni = unidirectional; FTF = face-to-face.
\end{flushleft}
\end{table}

Linear mixed-effects models were fit to each kinematic feature with fixed effects of Condition (back-to-back as the reference), Role (leader vs. follower), and trial order, with random intercepts for pair and individual nested within pair. At the global level, where features were averaged across the five-keypoint subset, both the unidirectional and face-to-face conditions were associated with greater movement magnitude than the back-to-back condition. In particular, subset velocity mean, velocity RMS, and displacement RMS were all significantly higher in both visually coupled conditions (see Table~\ref{tab:mirror_game_subset_lme}).

\begin{figure}
    \centering
    \includegraphics[width=\linewidth]{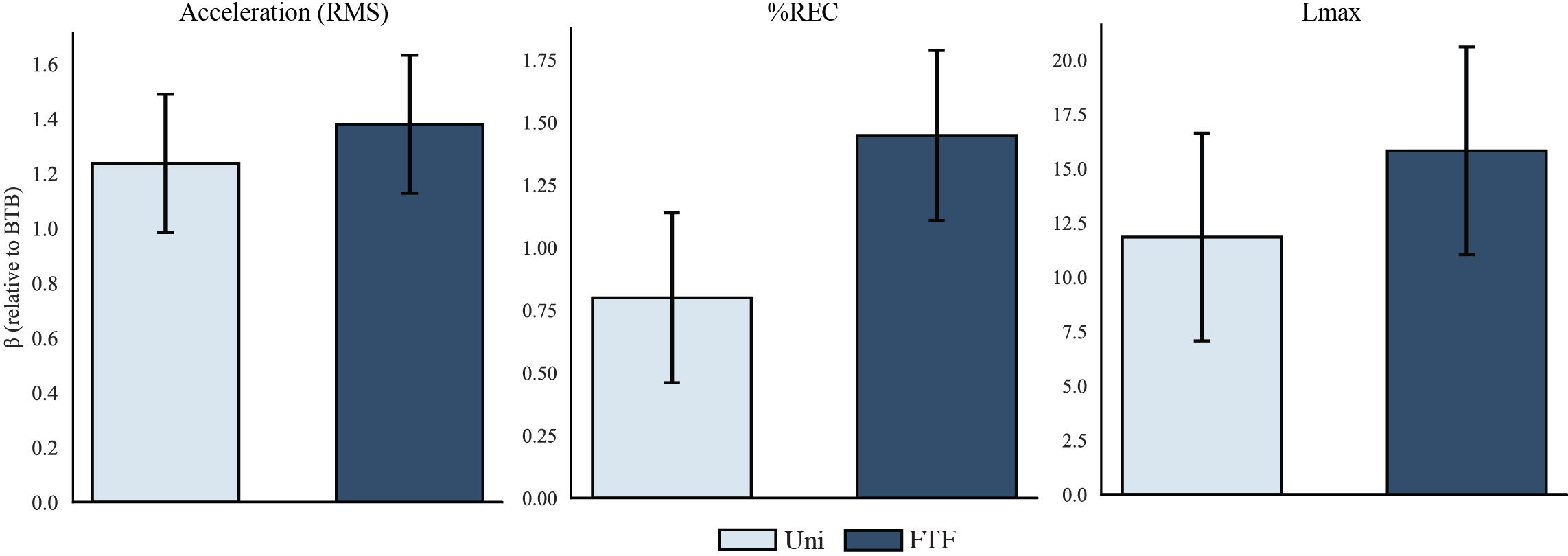}
    \caption{Effects of visual coupling on movement magnitude and dyadic coordination during the 3D Mirror Game. Bar plots display linear mixed-effects model coefficients ($\beta\pm SE$) for unidirectional (Uni) and face-to-face (FTF) conditions relative to the back-to-back (BTB) baseline. From left to right: acceleration (RMS), cross-recurrence rate (\%REC), and maximum diagonal line length (Lmax). All metrics were computed from a predefined five-keypoint subset (head, left/right wrists, left/right ankles) and averaged across keypoints prior to model fitting (see Methods for details).}
    \label{fig:mirror_game_results}
\end{figure}

The largest condition effects were observed for acceleration-based measures (Figure~\ref{fig:mirror_game_results}, left panel). Subset acceleration mean was significantly higher in both the unidirectional and face-to-face conditions relative to back-to-back, and subset acceleration RMS showed the largest condition coefficients overall. Across individual keypoints, kinematic features showed broadly similar condition-dependent increases, with the strongest effects generally observed at distal effectors such as the wrists and ankles. For example, left ankle and right wrist acceleration RMS were both elevated in the visually coupled conditions. In contrast, nose features were smaller overall, with reliable effects most consistently observed for acceleration-based measures and more pronounced increases in the face-to-face condition. Full results for each keypoint are reported in the supplementary materials. 

Together, these results show that increasing visual coupling systematically increased the overall movement and speed of whole-body motion, with the most pronounced differences emerging in acceleration-based metrics.

\paragraph{Cross-Recurrence Quantification Analysis}

Linear mixed-effects models were fit to each CRQA metric with fixed effects of Condition (back-to-back as reference), and Trial order, along with random intercepts for pair and individual-within-pair. CRQA metrics were averaged across the five keypoints for each trial before model fitting.  

Cross-recurrence rate increased significantly with visual coupling, indicating that leader-follower state-space trajectories visited similar regions more frequently when participants could see one another. Both face-to-face and unidirectional conditions showed elevated recurrence relative to back-to-back (FTF: $\beta = 1.45$, $SE = 0.34$, $p < .001$; uni: $\beta = 0.80$, $SE = 0.34$, $p = .019$). Despite this increase, determinism did not differ significantly across conditions (FTF: $\beta = -0.87$, $SE = 1.48$, $p = .555$; uni: $\beta = -1.52$, $SE = 1.48$, $p = .304$), nor did laminarity (FTF: $\beta = -0.014$, $SE = 0.011$, $p = .186$; uni: $\beta = -0.017$, $SE = 0.011$, $p = .117$).

The temporal characteristics of coordinated episodes, however, showed clear condition-dependent patterns. Maximum diagonal line length increased substantially with visual coupling (FTF: $\beta = 15.81$, $SE = 4.78$, $p = .001$; uni: $\beta = 11.85$, $SE = 4.78$, $p = .014$), indicating that periods of sustained coordination extended longer when visual information was available. Mean diagonal line length showed a similar pattern, but with only significant increases in the face-to-face condition (FTF: $\beta = 1.99$, $SE = 0.60$, $p = .001$; uni: $\beta = 0.81$, $SE = 0.60$, $p = .178$). The standard deviation of line lengths also increased in face-to-face interactions ($\beta = 2.79$, $SE = 0.89$, $p = .002$), reflecting greater variability in the duration of coordinated segments. Trapping time increased in the face-to-face condition ($\beta = 2.99$, $SE = 1.01$, $p = .003$), suggesting that mutual visual access enabled longer periods of quasi-stable coordination. These results are summarized in Figure~\ref{fig:mirror_game_results} (middle and right panels).

Entropy showed a marginal increase ($\beta = 0.13$, $SE = 0.08$, $p = .077$), while complexity showed a marginal decrease ($\beta = -0.14$, $SE = 0.08$, $p = .084$). These trends, though, did not reach significance, suggesting that visual coupling may have introduced greater diversity in the distribution of diagonal line lengths while simplifying the overall recurrence structure. Divergence, an index of system instability, did not vary significantly with condition.

Together, the CRQA results reveal that visual coupling enhanced the frequency and duration of coordinated states without altering the deterministic structure of the coupling. The kinematic results demonstrated that visual access increased overall movement magnitude and the CRQA findings extend this picture by showing that visual information also enabled longer sustained episodes of leader-follower synchronization.

\subsubsection{Discussion}

The mirror game case study demonstrates how the analysis pipeline extends to three-dimensional, dyadic coordination data while revealing complementary insights from linear kinematics and recurrence analysis. Linear kinematic features quantified how much participants moved, revealing a clear dose–response relationship in which visual coupling systematically increased movement magnitude and speed, with the strongest effects appearing in acceleration-based measures at distal effectors. This establishes that visual access enables more energetic, dynamic movement.

Recurrence quantification analysis extended these findings by addressing how the \textit{dyadic coordination evolved}. Cross-recurrence rate increased with visual coupling, confirming that leader-follower trajectories visited similar state-space regions more frequently when participants could see one another. The temporal structure of these coordinated episodes also changed with maximum diagonal line length nearly doubling in the face-to-face condition, indicating that visual feedback enabled longer periods of synchronization. Trapping time also increased, suggesting that mutual visual access permitted longer coordination states where one partner remained in a relatively fixed configuration while the other evolved. These signatures reveal that visual coupling not only increases how much people move but also alters the temporal architecture.

Principal component analysis provided an interpretable framework for characterizing whole-body movement structure. The first 14 principal movements captured approximately 96\% of postural variance, decomposing complex full-body kinematics into coherent modes of multijoint covariation. This data-driven decomposition parallels recent work by \citet{bigand_geometry_2024}, who used PCA to extract ``principal movements'' from naturalistic dance, revealing how different coordination modes synchronize through distinct spatial and temporal channels. The present analysis similarly reveals that PCA can decompose unconstrained dyadic coordination into interpretable synergies. 

These findings connect to the broader interpersonal coordination literature, which has established visual coupling as fundamental to interpersonal coordination \citep{gueugnon_postural_2016,noy_mirror_2011,hart_individuality_2014,zhai_design_2016}. Previous studies have predominantly used correlation-based metrics and computational modeling approaches to quantify coordination quality and individual differences \citep{hart_individuality_2014, noy_mirror_2011, zhai_design_2016}. The current results demonstrate the recurrence-based methods can provide complementary information. The flexibility found–increased recurrence rate without increases in determinism–suggest that adaptive coordination strategies emerge under rich visual information. This parallels findings that expert improvisation is characterized by more fluid adaptation rather than rigid synchronization \citep{hart_individuality_2014, noy_mirror_2011}. 

Methodologically, this case demonstrates the successful application of our analysis pipeline to full-body 3D data, extending earlier cases based on 2D facial and upper-body kinematics and highlighting the pipeline’s adaptability across technologies and scales. This consistency indicates that the pipeline offers a general framework suitable for diverse pose-estimation methods and research questions, addressing the field’s need for approaches that accommodate short time series, preserve temporal structure, and remain robust to non-stationarity. The successful scaling from fine-grained facial dynamics to whole-body interpersonal coordination demonstrates the approach’s potential to advance coordination science across multiple levels of analysis.

\section{General Discussion}\label{sec12}

This paper presented a general-purpose methodological framework for extracting meaningful patterns from human pose data across diverse experimental contexts. Through three case studies we demonstrated how the same methodological framework adapts to different pose estimation technologies, movement scales, and research questions. This pipeline's flexibility stems from its modular structure: pose acquisition, preprocessing, dimensionality reduction, and recurrence-based dynamics analyses can each be tailored to specific constraints while preserving the core analytic logic. 

The accessibility of the approach deserves emphasis. Case study 1 required only a consumer webcam, yet revealed load-dependent reorganization of facial dynamics and gaze-head decoupling that would have been invisible to traditional kinematic measures. This demonstrates that high-precision motion capture is not required when the goal is to characterize the temporal organization of behavior rather than fine-grained spatial accuracy \citep{seethapathi2019movement,roggio2024comprehensive}. In this sense, markerless pose estimation does not replace gold-standard systems but expands the range of settings in which meaningful dynamical structure can be measured. While cases 2 and 3 employed more sophisticated recording setups, the preprocessing and analysis steps remained conceptually identical, differing only in implementation details dictated by data dimensionality and noise characteristics \citep{nogueira2025markerless}.

Recurrence quantification analysis proved particularly valuable for revealing temporal structure that linear metrics miss \citep{marwan_recurrence_2007,webber2005recurrence}. Whereas linear approaches primarily capture changes in movement magnitude or average relationships under stationarity assumptions, RQA is sensitive to the organization of behavior in time, including regime shifts and transient structure. Across all three cases, RQA distinguished between quantitative changes in movement magnitude and qualitative reorganizations of coordination dynamics. The method captured state-dependent patterns---such as the fragmentation-then-reorganization trajectory under cognitive load, or the transition from rigid repetition to flexible recurrence with visual coupling–that conventional approaches would conflate \citep{riley1999recurrence,ramdani_recurrence_2013}. This sensitivity to nonlinear, nonstationary structure makes RQA well-suited to the inherent complexity of human movement \citep{bernstein1967coordination,stergiou2011human}.

Several practical lessons emerged from implementing the pipeline. First, missing data handling requires discipline: interpolation limits should be set conservatively rather than optimistically stretching to preserve sample size. Second, normalization decisions directly shape recurrence measures and must be chosen deliberately. Third, dimensionality reduction is not merely computational convenience but often essential for interpretability \citep{halilaj2018machine}. PCA revealed which movement patterns were most sensitive to visual information in case study 3, guiding interpretation in ways analyzing raw landmarks would not.

The pipeline's limitations warrant acknowledgment. Markerless pose estimation remains vulnerable to occlusion, lighting changes, and tracking errors, particularly for multi-person scenarios \citep{matsuda_validity_2024,evans_synchronised_2024}. While our preprocessing guidelines address common artifacts, they cannot eliminate all noise, and researchers must verify that observed effects exceed plausible artifact magnitudes. RQA parameter selection, though guided by principal methods like AMI and FNN, still involves judgment calls that can influence results. We recommend reporting preprocessing and parameter choices transparently and, where feasible, conducting sensitivity analyses to verify that conclusions hold across reasonable parameter ranges.

Computational demands increase steeply with data dimensionality and recording duration. MdRQA, while theoretically appealing for capturing collective dynamics \citep{wallot_multidimensional_2016}, quickly becomes intractable beyond a dozen or so features without aggressive dimensionality reduction. For very high-dimensional pose data or group coordination scenarios, researchers may need to balance comprehensiveness against computational feasibility, either by analyzing feature subsets separately or by accepting coarser temporal resolution.

Looking forward, several extensions merit exploration. Integrating pose with other modalities—such as speech, physiological signals, or eye tracking—could reveal cross-modal coordination patterns \citep{shockley2005cross}. Machine learning approaches for automated feature selection or parameter optimization could reduce the burden of manual tuning. Alternative state-space reconstruction methods beyond delay embedding might better suit certain movement types. Finally, standardized benchmarking datasets would facilitate systematic comparison of preprocessing and analysis choices across laboratories.

The proliferation of markerless pose estimation has democratized access to detailed movement data \citep{mathis2018deeplabcut,cao2019openpose}, but accessibility means little without principled analysis methods. This pipeline provides researchers with a coherent framework for moving from raw pose estimates to interpretable dynamical signatures, bridging computer vision capabilities with behavioral insight \citep{dindorf2024lab,ji2023review}. Whether the goal is monitoring cognitive state, quantifying interpersonal coordination, or characterizing movement disorders, the combination of careful preprocessing and recurrence-based analysis offers a practical path toward robust, reproducible pose-based behavioral research.

\backmatter

\section{Supplementary information}

Supplementary materials include full statistical results for case study 3, simulation details, formal definitions of recurrence metrics, and all analysis scripts, available at \\
\url{https://github.com/MInD-Laboratory/Pose-Dynamics}. The complete raw datasets for all three case studies can be accessed at \url{https://osf.io/bauqs}.

\bmhead{Acknowledgements}

We thank the following individuals for their contributions to the case studies reported in this manuscript: Rachel Kallen, Simon Hosking, Melissa Stolar, Michael Gostelow, Julia Wallier, Jan-Louis Kruger, and Mark Dras for their involvement in Case Study 1, and Brooke Luthy, Ronny Ibrahim, Jörg Buchholz, Craig Richardson, and Marcus Ockenden for their involvement in Case Study 2.

\section*{Declarations}

\begin{itemize}
\item Funding. Funders for this project include the Australian Future Hearing Initiative, a collaboration formed under Google’s Digital Future Initiative in partnership with Macquarie University, and the Defence Science and Technology Group (DSTG), Australian Department of Defence, via the Centre for Advanced Defence Research in Robotics and Autonomous Systems (CADR-RAS). MCM was supported by a Macquarie University Lighthouse Fellowship. 
\item Competing interests. The authors have no relevant financial or non-financial interests to disclose.
\item Ethics approval and consent to participate. All experimental protocols were approved by the Macquarie University Human Research Ethics Committee. Informed consent was obtained from all participants prior to participation.
\item Consent to participate. Not applicable
\item Consent for publication. Not applicable
\item Data availability. The complete raw dataset can be accessed at \url{https://osf.io/bauqs}
\item Code availability. All analysis scripts and results are available at \url{https://github.com/MInD-Laboratory/Pose-Dynamics}. 
\item Author contributions. C.S., M.C.M., G.P., S.W., R.W.K, and M.J.R. conceived the study. C.S., M.C.M., G.P., M.J.R., and K.M. developed the methodology and implemented the analyses. C.S., M.C.M., G.P., and K.M. contributed to data collection and case study preparation. C.S. and M.C.M. wrote the first draft of the manuscript. All authors contributed to manuscript revision and approved the final version. M.J.R. supervised the project.
\end{itemize}

%\noindent
%If any of the sections are not relevant to your manuscript, please include the heading and write `Not applicable' for that section. 

\newpage
\bibliography{sn-bibliography}

\end{document}